\documentclass[12pt,letterpaper]{article} 
\usepackage[top=0.85in,left=2.75in,footskip=0.75in]{geometry}

\usepackage{amsmath,amssymb}

\usepackage{changepage}

\usepackage{textcomp,marvosym}

\usepackage{cite}

\usepackage{nameref,hyperref}

\usepackage[right]{lineno}

\usepackage{microtype}
\DisableLigatures[f]{encoding = *, family = * }

\usepackage[table]{xcolor}

\usepackage{array}

\newcolumntype{+}{!{\vrule width 2pt}}

\newlength\savedwidth

\newcommand\thickhline{\noalign{\global\savedwidth\arrayrulewidth\global\arrayrulewidth 2pt}%
\hline
\noalign{\global\arrayrulewidth\savedwidth}}

\raggedright
\usepackage[aboveskip=1pt,labelfont=bf,labelsep=period,justification=raggedright,singlelinecheck=off]{caption}

\makeatletter
\renewcommand{\@biblabel}[1]{\quad#1.}
\makeatother

\usepackage{lastpage,fancyhdr,graphicx}
\usepackage{epstopdf}
\fancyheadoffset[L]{2.25in}
\fancyfootoffset[L]{2.25in}
\usepackage[table]{xcolor}

\usepackage{array}

\usepackage{algorithm}
\usepackage[noend]{algpseudocode}
\algrenewcommand\textproc{}

\usepackage[finalnew]{trackchanges}

\usepackage{subfig}

\addeditor{RG}
\addeditor{LK}
\addeditor{m} 
\addeditor{R1}
\addeditor{R2}
\addeditor{R3}

\def\appref#1{Appendix~\ref{#1}}

\long\def\comment#1{}
\def\eg{{\em e.g.},\ }
\def\ie{{\em i.e.},\ }
\def\etal{{\em et~al.}}

\def\CPR#1#2{\PR{#1\,|\,#2}}

\def\KM#1{\hbox{KM}(\, #1\,)}
\def\KMinv#1{\hbox{KM}^{-1}(\, #1\,)}
\def\PSSP#1#2#3{  
  Pr_{#1}(\,#2 \ |\ #3\,) }

\def\pat{{\bf x}} 

\def\mm{$-$}
\def\pp{$+$}

\def\strain{c\_topic}
\def\AlgName#1{{\sc #1}}
\def\PreProcess#1{\AlgName{PreProcess$_{#1}$}}

\def\gev#1#2{e_{#1}^{#2}}  

\def\CPRsub#1#2#3{P_{#1}(\  #2\ |\ #3\ )}
\def\CPR#1#2{\CPRsub{}{#1}{#2}} 

\def\LSM{{\sc LSM}} 
\def\USM{{\sc USM}} 
\def\ComputeBasis{{\sc ComputeBasis}}
\def\UseBasis{{\sc UseBasis}}
\def\ComputedLDA{{\sc Compute}\_d{\sc LDA}}
\def\UsedLDA{{\sc Use\_dLDA}}

\def\UseModel{{\sc UseModel}}
\def\tech#1{Enc\_#1}  

\def\ParamAlg#1#2#3{{\sc #1}{\scriptsize #2}#3}

\def\Lbl{Lbl}
\def\docParam{\theta}  
\def\wordParam{\beta} 
\def\wordProb{\bar{\beta}} 
\def\tpc#1{\wordProb^{#1}}   
\def\dcm#1{f_{#1}}   

\def\sCox{Cox\hspace*{7em}}
\def\sRCox{\hspace*{3em}RCox\hspace*{4em}}
\def\sMTLR{\hspace*{6em} MTLR}

\def\tpcD#1#2{\docParam_{#1}(\,\dcm{#2}\,)}
\def\dtd#1{\Theta(\,\dcm{#1}\,)} 

\begin{document}
\vspace*{0.2in}

\begin{flushleft}
{\Large
\textbf\newline{Gene Expression based Survival Prediction for Cancer Patients -- A Topic Modeling Approach} 
}
\newline
\\
Luke Kumar\textsuperscript{1,2\Yinyang},
Russell Greiner\textsuperscript{1,2\Yinyang},
\\
\bigskip
\textbf{1} Department of Computing Science, University of Alberta, Edmonton, Alberta, Canada
\\
\textbf{2} Alberta Machine Intelligence Institute (Amii), Edmonton, Alberta, Canada
\\
\bigskip

%
%
\Yinyang These authors contributed equally to this work.





* rgreiner@ualberta.ca

\end{flushleft}
\section*{Abstract}
Cancer is one of the leading cause of death, worldwide.
Many believe that genomic data will enable us to
better 
predict the survival time
of these patients,
which will lead to better, more personalized treatment options and patient care.
As standard survival prediction models 
have a hard time coping 
with the high-\-dimensionality of such gene expression data,
many projects use some dimensionality reduction techniques to overcome this hurdle. 
We introduce a novel methodology,
inspired by topic modeling from the natural language domain,
to derive expressive features from the high-dimensional gene expression data. 
There, a document is represented as a mixture over a relatively small number of topics,
where each topic 
corresponds to 
a distribution over the words;
here, to accommodate the heterogeneity of a patient's cancer,
we represent each patient ($\approx$~document) as a mixture over cancer-topics, where each cancer-topic is a mixture over gene expression values ($\approx$~words).
This required some extensions to the standard LDA model
-- \eg to accommodate the {\em real-valued}\ expression values -- leading to
our novel {\em discretized Latent Dirichlet Allocation}\
(dLDA) procedure.
After using this dLDA
to learn these cancer-topics,
we can then express each patient as a distribution over a small number of cancer-topics,
then use this low-dimensional
``distribution vector'' 
as input to a learning algorithm --
here, we ran the recent survival prediction algorithm, MTLR, 
on this representation of the cancer dataset. 
We initially 
focus on the METABRIC dataset,
which describes each of n=1,981 breast cancer patients
using the r=49,576 gene expression values, from microarrays.
Our results show that our approach (dLDA followed by MTLR) provides survival estimates
that are more accurate than standard models, 
in terms of the standard Concordance measure.
We then validate this ``dLDA+MTLR'' approach 
by running it on the n=883 Pan-kidney (KIPAN) dataset,
over
r=15,529 gene expression values --
here using the mRNAseq modality 
--
and find that it again achieves excellent results.
In both cases, we also show that the resulting model is calibrated, 
using the recent ``D-calibrated'' measure.
These successes, in two different cancer types and 
expression modalities, 
demonstrates the generality,
and the effectiveness, of this approach.

\section{Introduction} 
\label{sec:Intro}
The World Health Organization reports that cancer has become the second leading cause 
of death globally,
as approximately 1 in 6 deaths are caused by some form of cancer~\cite{stewart2017world}.
Moreover,
cancers are very heterogeneous,
in that the outcomes can vary widely for patients with similar diagnoses, 
who receive the same treatment regimen.
This has motivated researchers to seek other features
to help predict individual outcomes.
Many such analyses use just clinical features.
Unfortunately, features such as lymph node status and histological grade, 
while 
predictive of metastases, 
do not appear to be sufficient to reliably categorize clinical outcome~\cite{van2002gene}.
This has led to many efforts to improve the prognosis for
cancer, based on genomics data 
(\eg gene expression (GE) or copy number variation (CNV) ),
possibly
along with the clinical data~\cite{margolin2013systematic, parker2009supervised, naderi2007gene, van2002gene,beer2002gene}.
Focusing for now on breast cancer, 
van't~Veer~\etal~\cite{van2002gene} used the expression of 70 genes to 
distinguish  high vs low risk of distant metastases 
within five years.
Parker~\etal~\cite{parker2009supervised} identified five subtypes of breast cancer,
based on a panel of 50 genes (PAM50): 
luminal A, luminal B, HER2-enriched, basal-like, and normal-like.
Later, Curtis~\etal~\cite{curtis2012genomic} examined $\approx$2000 patients from a wide study combining clinical and genomic data, 
and identified around ten subtypes.
All three of these studies showed that their respective subtypes
produce significantly different Kaplan-Meier survival curves~\cite{kaplan-meier},
suggesting such molecular variation does influence the disease progression.%
\footnote{
There are also many other systems that use such expression information to 
divide the patients into two categories: high- vs low-risk; 
{\em cf.}, \cite{beer2002gene,van2002gene}.
}

More recently, many survival {\em prediction}\ models have been
applied to cancer cohorts, 
with the goal of estimating survival times for individual patients; 
some are based on standard statistical survival analysis techniques, and others based on classic regression algorithms -- 
\eg random survival forests~\cite{RandomSurvivalForests}
or \add[R1]{support vector regression for censored data (SVRc)\hbox{~\cite{khan2008support}}}.
With the growing number of gene expression experiments being cataloged for analysis, 
we need to develop survival prediction models that can utilize such high dimensional data.
Our work describes such a system
that can learn effective survival prediction models
from high-dimensional gene expression data.

The 2012 DREAM Breast Cancer Challenge (BCC) was designed
to focus the community's efforts to improve breast cancer survival prediction~\cite{margolin2013systematic}.
Its organizers made available clinical and genomic data (GE and CNV) of $\approx$2000 patients from the 
\cite{curtis2012genomic} study (mentioned above).
Each submission  to the BCC challenge 
mapped each patient to a single real value (called ``risk''),
which is predicting that patients with higher risk should die earlier than those with lower risk.
The entries were therefore evaluated based on the concordance measure:
basically, the percentage of these pairwise predictions that were correct~\cite{kalbfleisch2011statistical}.
This is standard, 
in that many survival prediction tasks use the concordance as the primary measure to assess the performance of the survival predictors,
here and in other challenges%
~\cite{guinney2016prediction}.
The winning model~\cite{cheng2013development} performed statistically better than the state-of-the-art benchmark models~\cite{margolin2013systematic}.

This paper explores several dimensionality reduction technique,
including a novel approach based on topic modeling,
``discretized Latent Dirichlet Allocation'' (dLDA),
seeking one that can produce highly predictive 
 features from the high-dimensional gene expression data.
We explored several ways to apply this 
topic-modeling approach to gene expression data,
to identify the best ways to use it to map the gene expression description into a much lower dimensional description 
(from $\approx$50K features to 30 in this METABRIC dataset).
We then gave the resulting transformed data as input to  a recently-developed non-parametric learning algorithm, multi-task logistic regression (MTLR), which produced a model that can then predict an individual's survival distribution~\cite{Yu_al_NIPS11}.
We show that this predictor performs better than other standard survival analysis tools
in terms of concordance.
We also found that it was ``D-calibrated''~\cite{AndresPLoSONE18,ISD-Paper}; see
Appendix~\ref{app:d-calib}.

To test the generality of our learning~approach (dLDA + MTLR),
we then applied the same learning algorithm 
-- the one that worked for the METABRIC microarray gene expression dataset --
to the Pan-Kidney dataset, 
which is a different type of cancer (kidney, not breast),
and is described using a different type of features
(mRNAseq, not microarray).
We found that the resulting predictor was also extremely effective,
in terms of both concordance and D-calibration.

This paper provides the following three contributions: 
(1)~We produce an extension to LDA, called ``dLDA'', needed to handle continuous data;
(2)~we use this as input to a survival prediction tool, MTLR -- 
introducing that tool to this bioinformatics community;
and (3)~we demonstrate that this dLDA+MTLR combination works robustly, 
in two different datasets, using two different modalities
-- working better than some other standard approaches, 
in survival prediction.

Section~\ref{sec:Foundation} 
introduces the basic concepts, related to the survival prediction task in general
and latent dirichlet allocation;
Section~\ref{sec:Data} then describes the datasets used in this study; 
and
Section~\ref{sec:overview_LP} presents an overview of learning and performance tasks, 
at a high level.
Section~\ref{sec:results} 
(resp., \ref{sec:discussion}, \ref{sec:conclusion})
then presents our results (resp., discussions, contributions).
The supplementary appendices provide additional
figures, tables, and other and
material
-- \eg defining some of the terms, and introducing ``D-calibration''.

\section{Foundations} 
\label{sec:Foundation}
This section  
provides the foundations:
Section~\ref{sec:SurvivialPrediction}
overviews 
the survival prediction task in general
then Section~\ref{sec:dLDA} 
describes  
Latent Dirichlet Allocation (LDA),
first showing its original natural language context, 
then discussing how 
we need 
to extend it for our gene expression context.
These significant modifications
lead to a discretized variant, dLDA.
We also contrast this approach with other survival analysis of gene expressions.

\subsection{Survival prediction} 
\label{sec:SurvivialPrediction} 

\begin{figure}[t] \centering
\includegraphics[scale=0.5]{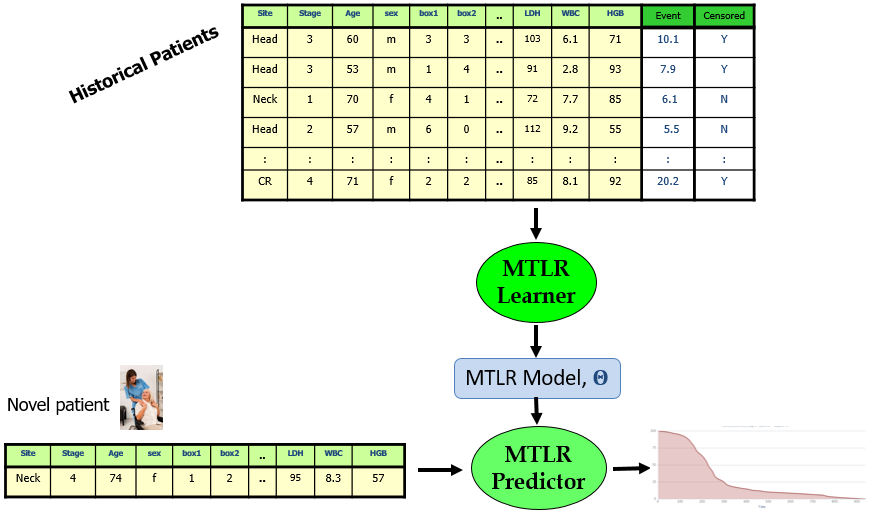}
\caption{Survival prediction training and performance tasks. 
{\it Training-Task:} Historical data with event times and censor-status along with covariates are used to train a model -- top-to-bottom. 
{\it Performance-Task:} new patient covariates are input to the learned model to produce a prediction of survival time -- bottom, left-to-right.
(Included picture designed by Freepik.)}
\label{fig:ISD_model}
\end{figure}

Survival prediction is similar to regression 
as both involve learning a model that regresses the covariates 
of an individual to 
estimate the value of a dependent real-valued response variable
-- here, that variable is ``time to event'' (where the standard event is ``death'').
But survival prediction differs from the standard regression task as 
its response variable is not fully observed in all training instances
-- this tasks allows
many of the instances 
to be ``right censored'',
in that we only see a {\em lower bound}\ of the response value.
This might happen if a subject was alive when the study ended,
meaning we only know that she lived {\em at least}\ (say) 5 years
after the starting time, 
but do not know whether she actually lived 5 years and a day, or 30 years.
This also happens if a subject drops out of a study, after say 2.3 years, and is then lost to follow-up; etc.
Moreover, one cannot simply ignore such instances as it is common for many (or often, {\em most}) of the training instances to be right-censored; see Table~\ref{tab:data}.
Such ``partial label information'' is problematic for standard regression techniques,
which assume the label is completely specified for each training instance.
Fortunately, there are survival prediction algorithms that can learn an effective model, 
from a cohort that includes such censored data.
Each such dataset contains descriptions of a set of instances (\eg patients),
as well as two ``labels'' for each:
one is the time, 
corresponding to the 
{\em time from diagnosis to a final date} (either death, or time of last follow-up) and
the other is the {\em  status} bit,
which indicates whether the patient was alive at that final date (Figure~\ref{fig:ISD_model}). 

\subsubsection{Patient Specific Survival Prediction 
\change[m]{via}{using the} MTLR Model}
\label{sec:MTLR}
\begin{figure}\centering
\includegraphics[width =5in,height=2in]{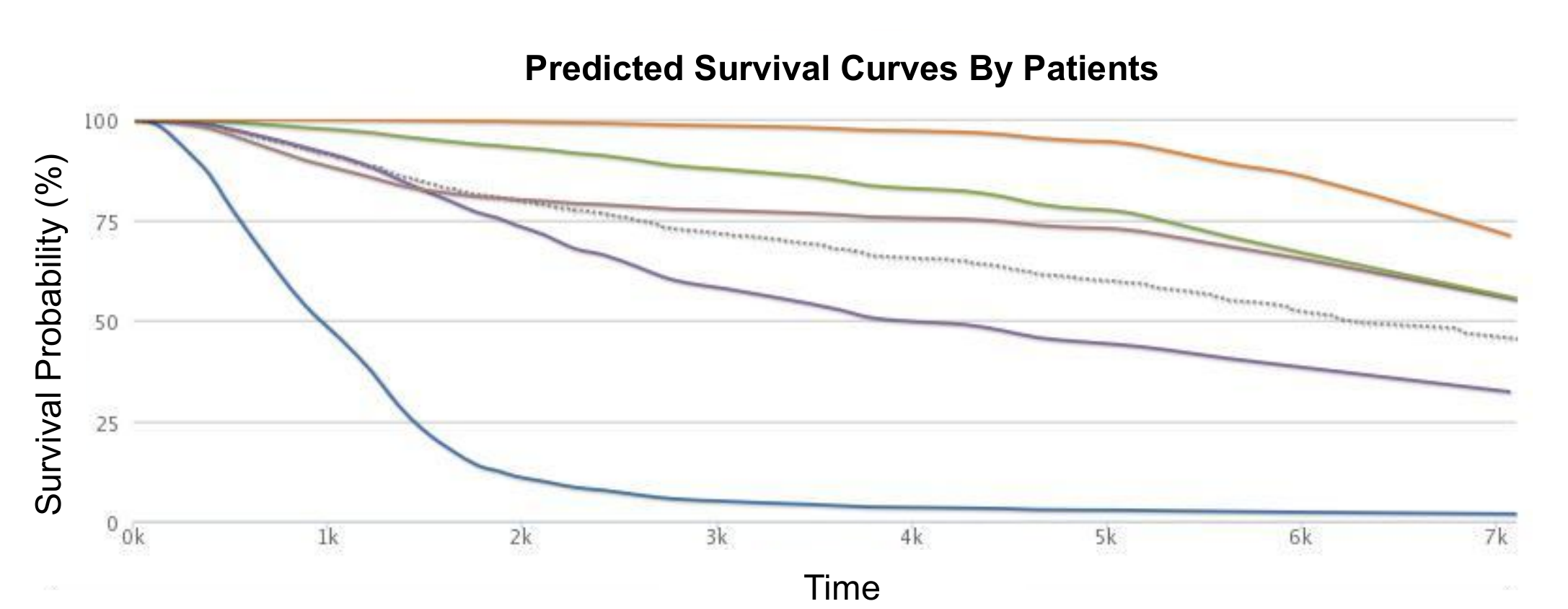}
\caption{
\note[R3]{We rewrote this caption, and also modified the figure itself.}
The dashed line is the Kaplan-Meier plot for this dataset.
Each of the other 5 curves is a patient-specific survival curve, for 5 different METABRIC patients,
from a learned MTLR model.
The curves show that there are 
very different prognoses for the different patients,
even though they all are breast cancer patients from the same cohort: 
here, 
the patient with the orange survival curve 
(near the top) has a very good prognosis,
especially compared to the patient with the blue survival curve.
}
\label{fig:pssp_surv}
\end{figure}

This project considered 3 ways to learn a survival model:
The standard approaches
-- Cox and \change[R1]{regularized RCox}{Regularized Cox (RCox)} --
are overviewed in Appendix~\ref{sup:cox}.
This subsection describes the relatively-new 
MTLR~\cite{Yu_al_NIPS11} 
system, which learns a model (from survival data) that,
given a description of a patient $\pat \in \Re^r$,
produces a 
{\em survival}\ curve, 
which specifies 
the probability of death $D$, $\CPR{D \geq t}{\pat}$ vs $t$
for all times $t\geq 0$.
This survival curve is similar to a Kaplan--Meier curve~\cite{kaplan-meier}, but incorporates all of the patient specific features $\pat$.
In more detail:
MTLR first identifies $m$ time points $\{ t_i \}_{i=1..m}$
and then learns a variant of a logistic regression function, parameterized by 
$W = \{[w_i, b_i]\}_{i=1..m}$ 
over these $m$ time points,
a different such function for each time $t_i$ -- 
meaning $W$ is a matrix of size $m \times (r+1)$.
Using the random variable $D$ for the time of death 
for the patient described by $\pat$: 
\begin{equation}\small
\PSSP{W}{D \in [t_{k}, t_{k+1})}{\pat} 
 \quad \propto \quad
\exp\left( \sum_{\ell={k+1}}^m ({w_\ell}^T \pat \ +\ b_\ell)\,
\right) 
\label{eqn:PSSP}
\end{equation}
The MTLR model then combines 
\change[m]{these PMFs}{the values of the PMF} (probability mass function) into a CMF (cumulative mass function),
by adding them in the reverse order --
hence
from the probability value of $1$ at $t_0=0$ 
-- \ie $\PSSP{W}{D \geq 0}{\pat}
\ =\ 1$
-- down to smaller values as the time $t$ increases. 
Figure~\ref{fig:pssp_surv} shows the individual survival curves of several patients.
Here, we view a patient's descretized survival time $d \in \Re^{\geq 0}$ as a binary vector (of classification labels) 
$y(d)\ =\ [y_1(d),\ y_2(d),\ \dots,\ y_m(d)]$,
where each $y_j(d)\in \{0,1\}$ encodes that patient's survival status at each time interval  $[t_j,\, t_{j+1}]$:
$y_j(d)=0$ (no death yet) for all $j$ with $t_j < d$ and $y_j(d)=1$ (death) for all $t_j \geq d$.
The learning system attempts to optimize 

{ \small
\begin{equation}
\begin{array}{l}
\displaystyle \min_{W} \ \frac{C}{2} \sum_{j=1}^{m} \|w_j \|^2 
\quad - 
 \quad \displaystyle \sum_{i=1}^{n} \left [ \sum_{j=1}^{m} y_j(d_i)(w^T \pat_i + b_j) - \log \sum_{k=0}^{m} \exp (f_{W}(\pat_i,k)) \right]\label{eq_5} \\
\hbox{where}\quad \displaystyle f_{W}( \pat_i,k) \quad=\quad 
\sum_{\ell=(k+1)}^{m} (w_\ell^T \pat_i + b_\ell) 
\qquad \hbox{for} \ \ 0 \leq k \leq m \\[-1ex]
\end{array}
\end{equation} }

\noindent
(This formula applies to uncensored patients;
we apply the obvious extension to deal with censored instances.)
This overall equation includes a L2 regularization term
to reduce the risk of overfitting. 
The MTLR parameter $m$ (the number of time points) is set to the square-root of the number of instances in all our experiments.

Given the learned parameters $W$, 
we can then use Equation~\ref{eqn:PSSP}
to produce a curve for each patient;
we can then use the (negative of) the mean of the patient's specific predicted survival distribution
as her risk score.
Yu~\etal~\cite{Yu_al_NIPS11} presents more detailed explanations of model formulation, parameter learning ($W$), and the prediction task.
MTLR differs from many other models (such as the standard Cox model)
as: 
(1)~MTLR produces a survival function, rather than just a risk score; and
(2)~MTLR does not make the proportional hazards assumption -- 
\ie
it allows effect of each covariate to change with time.
See also Haider~\etal~\cite{ISD-Paper}.
Note this is the learning process of 
\add[R1]{LearnSurvivalModel} (\ParamAlg{LSM}{[$\Psi\,=\,$MTLR]}{}) 
appearing below in Figure~\ref{fig:LSM}, 
and Section~\ref{sec:LSM}.

\subsection{Discretized Latent Dirichlet Allocation (dLDA)}  
\label{sec:dLDA}

Latent Dirichlet Allocation (LDA) is a widely used generative model~\cite{blei2003latent}, 
with many successful applications in natural language (NL) processing. 
LDA views each document as a distribution over multiple topics (document-topics distribution), where each topic is a distribution over a set of words (topic-words distribution)
-- that is, LDA assumes that each word in a document is generated by first sampling a topic from the document's document-topics distribution and then sampling a word from the selected topic's topic-words distribution.
Given the set of topics 
(each corresponding to a specific topic-word distribution), 
we can view each document
 as its distribution over topics, which is very low dimensional.

The LDA {\em learning process}\ first identifies the latent topics
-- that is, the topic-words distributions corresponding to each latent topic --
based on the words that frequently co-occur across multiple documents;
{\em n.b.}, it just uses the documents themselves,
but not the labels.
For example, it might find that many documents with the word ``ball'' 
also included ``opponent'' and ``score''; and vice versa.  
Similarly, ``finances'', ``transaction'', and ``bank'' often co-occur,
as do ``saint'', ``belief'' and ``pray''.
Speaking loosely,
the topic-model-learner might then form one topic, $\tpc{1}$, that gives high probabilities to the first set of words (and relatively low probabilities to the remaining words) -- 
perhaps~\footnote{
Technically, each topic is specified as a Dirichlet distribution over the set of words, $\wordParam^i$.
To simplify the presentation,
here we are showing their expected values,
these $\wordProb^i = E[ \wordParam^i ]$ values are 
based on the priors;
see Appendix~\ref{sec:ComputeBasis}(3).
}
$$\begin{array}{lcl}
\CPR{\hbox{ball}}{\tpc{1}} & =& 0.05\\ 
\CPR{\hbox{opponent}}{\tpc{1}}& =& 0.03\\
\CPR{\hbox{score}}{\tpc{1}} & =& 0.01\\
\CPR{x}{\tpc{1}} & < &\hbox{1E-4 \quad for all other words\ } x
\end{array}$$

This $\tpc{1}$ corresponds to an $n$-tuple over the $n$ words; 
we call this $\tpc{1} \approx  [\CPR{w_1}{\tpc{1}},\ \dots,\ \CPR{w_n}{\tpc{1}}]$.
It would similarly identify a second topic $\tpc{2}$
with the $n$-tuple 
$\tpc{2} \approx  [\CPR{w_1}{\tpc{2}},\ \dots,\ \CPR{w_n}{\tpc{2}}]$
that gives high probabilities to the different set of words, etc.
(While we might view the first topic as related to sports, 
the second related to finances, and third to religion,
that is simply our interpretation, and is not needed by the learning algorithm.
Other topics might not be so obvious to interpret.)
This produces the {\em topic-words distribution} 
${\cal B}\ =\ \{ \tpc{i} \}_{i=1..K}$ over $K$ topics.

The learner would then map each document into a ``distribution'' 
over this set of $K$ topics -- perhaps document
$\dcm{1}$ would be decomposed as
$\dtd{1}\ =\ 
[\tpcD{1}{1},\, \tpcD{2}{1},\, 
\dots,\, \tpcD{K}{1}]\ =\ $[0.01, 13.02, 50.01
\dots, 0.03]
-- these are parameters for a Dirichlet distribution,
which are non-negative, but do not add up to 1.
These are different for different documents
-- \eg 
\add[m]{perhaps}
$\dcm{2}$ is expressed as 
$\dtd{2}\ =\ 
$[12.03, 0.001, 3.1, \dots, 2.4], etc.
This is the {\em document-topic distribution} $\{\, \dtd{j}\, \}_{j=1..m}$ over the $m$ documents.

The specific learning process depends on the distributional form of the document-topics and topic-words distributions
(here, we use Dirichlet for both) 
and also the number of latent topics, $K$.
Given this, the LDA learning process
finds the inherent structure present in the data
-- 
\ie a model (topic-words distributions for each of the $K$ topics 
$\{ \tpc{i} \}_{i=1..K}$) 
that maximizes the likelihood of the training data.

The same way certain sets of words often co-occur in a document, 
similarly sets of genes are known to be co-regulated:
under some condition 
(corresponding to a ``\strain''),
every gene in that set will have some additional regulation -- 
some will be over-expressed,
each by its own amount,
and the others will be under-expressed.
Moreover, just as  
\change[R1]{an NL}{a natural language (NL)} topic typically involves
relatively few words,
most \strain{}s effectively involve relatively few genes.
Also, just like a document may involve a mixture of many topics,
each to its own degree,
so a patient's cancer often involves multiple
\strain{}s;
see work on cancer subclones~\cite{deshwar2015phylowgs}.
This has motivated many researchers to use
some version of topic modeling to model
gene expression values,
under various (sets of) conditions.

For example, Rogers~\etal~\cite{rogers2005latent} proposed 

 Latent Process Decomposition (LPD),
a probabilistic graphical model
that was inspired by LDA, for microarray data,
and presented clustering of genes 
that led to results comparable to those produced by hierarchical clustering.
(Their results are descriptive;
they do not use the results in any downstream evaluation.) 
\add[R1]{Later Masada~\hbox{\etal~\cite{masada2009bayesian}} proposed improvements to the original LPD approach and showed similar results.}
Bicego~\etal~\cite{bicego2012investigating} report topic modeling approaches (including LPD) were useful in classification tasks with gene expression data.
They applied several topic models as dimensionality reduction tools to 10 different gene expression data sets,
and found that the features from the topic models led to better
predictors.

\add[R1]{Further,
Lin~\hbox{\etal~\cite{liu2016overview}}
reviewed various different topic models applied to  gene expression data,
including LDA and probabilistic latent semantic analysis (PLSA)~\hbox{\cite{hofmann2001unsupervised}},
as well as 
the topic model approaches described above, for gene classification and clustering.
They note that the topic model approaches improve over other models
as one can easily interpret the topic-words distributions and the mixed membership nature of the document-topic distribution.
} 

However, none of these tasks were survival analysis.
They used shifting and scaling to convert the continuous gene expression values to discrete values;
we considered this approach for our data,
but found that it was not able to learn distinct topics for our data. 
Moreover, 
\change[m]{all patients had}{this gave all patients} very similar document-topic distributions. 
Dawson~\etal~\cite{dawson2012survival} proposed a survival supervised LDA model, called survLDA,
as an extension of supervised LDA~\cite{mcauliffe2008supervised}. 
survLDA uses a Cox model~\cite{coxmodel} to model
the response variable (survival time) 
instead of the generalized linear model~\cite{mccullagh1989generalized} 
used in supervised LDA~\cite{mcauliffe2008supervised}. 
But 
Dawson~\etal~\cite{dawson2012survival} reported 
\remove[m]{in their empirical finding}
that the topics learned from survLDA 
were very similar to the ones learned from the general (unsupervised) LDA model.

Here, we apply the 
``standard'' topic-modeling approach
to gene expression data, for the survival prediction task. 
While previous systems applied topic modeling techniques to gene expression data,
very few have applied topic models to predict a patient's survival times 
(and none to our knowledge  have 
used mRNAseq expression data).
Our work presents a more direct analogue to the NL topic modeling that can be applied to our cohort of patients with gene expression data,
where each patient 
corresponds to a document and the genes/probes in the expression data correspond to the words that form the document.
This requires making some significant modifications to the standard LDA model,
which assumes  
the observations are frequencies of words,
which are non-negative integers that generally follows a monotonically decreasing distribution.
By contrast, gene expression values are arbitrary real values, 
believed to follow a skewed Gaussian distribution~\cite{wolfinger2001assessing};
see also Figure~\ref{fig:encoding_AB}. 
{(This is also true for mRNAseq, as we need to normalize the expression counts to be comparable, from patient to patient.)}

We follow 
the approach 
of explicitly discretizing the expression values
in a preprocessing step,
so the resulting values
basically, approximate a Zipf distribution. 
There are still some subtleties here
-- \eg while the NL situation involves only non-negative integers,
an affected gene can be either over-expressed, or under-expressed
-- \ie we need to deal with two directions of ``deviation'', 
while NL's LDA just deals with one direction;
see Section~\ref{sec:ComputeBasis-dLDA}.
We refer to our model as {\em dLDA}\ and the discretized gene expression values as {\em dGEVs}.
The same way the standard LDA approach reduces
the description of a document from 
a $\approx 10^5$-dimensional vector (corresponding to the words used in that document)
to a few dozen values
(the ``distribution'' of the topics),
this dLDA approach reduces the $\approx 50K$-dimensional gene expression tuple
to a few dozen values
-- here the ``distribution'' of the \strain{}s.
Figure~\ref{fig:C-U-Basis} summarizes this process:
using the subroutines defined in Section~\ref{sec:overview_LP}
below,
at learning time,
\ParamAlg{ComputeBasis}{[$\rho=$dLDA]}{}\ 
first identifies the set of relevant \strain{}s $\wordProb_{GE}$
from the set of gene expression values $X'_{GE}$,
then later (at performance time), 
\ParamAlg{UseBasis}{[$\rho=$dLDA]}{}\ 
uses those learned \strain{}s
to transform a new patient's high-dimensional gene expression profile 
$x'_{GE}$ to a low-dimensional \strain-profile, $x"_{GE}$
-- here going from 50K values to 30.

Section~\ref{sec:results} presents empirical evidence
that this method works effectively for our survival prediction task;
\appref{sup:lpd}
shows that it  
performs better than the LPD technique.

\section{Datasets Used} 
\label{sec:Data}
\begin{table} 
\caption{Characteristics of METABRIC and KIPAN Cohorts
\label{tab:data} }
\small
\begin{tabular}{@{}l|l|l@{}}
\hline 
\rowcolor{gray!20}  & METABRIC & KIPAN \\ 
\thickhline
\# Patients$^a$  &  1,981 & 883 \\
\# Censored & 1,358 ($\sim 68.5$\%) & 655 ($\sim 74.4$\%) \\
\# Uncensored & 623 & 228 \\
Time span in days (Uncensored) & 3 -- 8,941 & 2 -- 5,925 \\
\# Clinical features & 19 & 10 \\
\# Expressions ($\approx$ \#genes)$^b$ & 49,576 (probes) & 15,529 \\
Gender &  Women (100\%) & Women (32.7\%), Men (67.3\%) \\
\hline
\end{tabular}
\\
{\small
$^a$~We removed the $\sim$50 patients from the KIPAN dataset
that did not contain mRNAseq data.\\
$^b$~While METABRIC also included copy number variations (CNV) data for the patients, 
here we focus on only gene expression data.
}
\end{table}

We apply our methods to two large gene expression
datasets: 
the METABRIC breast cancer cohort~\cite{curtis2012genomic} (mircroarray) and 
the Pan-kidney cohort KIPAN (mRNAseq)~\cite{kipan}
\footnote{
\add[R3]{http://firebrowse.org/?cohort=KIPAN\&download$\_$dialog=true
}}.
We initially focus on the METABRIC dataset~\footnote{
\add[R3]{https://www.synapse.org/\#!Synapse:syn1688369/wiki/27311}},
which is one of the largest available survival studies
that includes genomic information.
In 2012, the Breast Cancer Prognostic Challenge (BCC) organizers released 
the METABRIC (Molecular Taxonomy of Breast Cancer International Consortium) dataset
for training~\cite{curtis2012genomic}.
While they subsequently released a second dataset (OSLO)
for final testing~\cite{curtis2012genomic},
we are not using it for several reasons:
(1)~METABRIC provided disease-specific survival (DS), 
which considers only {\em breast cancer death}\ (BC-based death),
rather than all causes of death~\cite{cheng2013development}.
By contrast, OSLO provides ``overall survival'', 
which does not distinguish BC-based deaths from others.
As DS is clearly better for our purpose, it is better to evaluate on the METABRIC dataset.
(2a)~OSLO and METABRIC contained different sets of probes
-- and in particular, OSLO contains only $\sim80$\% of the METABRIC probes.
(2b)~Similarly, the OSLO dataset is also missing some of the clinical covariates that are present in the METABRIC dataset 
-- \eg menopausal status, group, stage, lymph nodes removed, etc.; 
see \cite[Table~1]{margolin2013systematic}.
This means a ``METABRIC-OSLO study'' would need to exclude some METABRIC features and some METABRIC probes.

We then used a second independent dataset,
to verify the effectiveness of our ``dLDA+MTLR'' approach.
Here, we did not use OSLO, 
as we wanted to explore a different type of cancer,
and also use a different platform,
to show that our system could still identify an appropriate 
(and necessarily different) set of cancer-topics (\strain{}s).
We therefore used the KIPAN dataset from TCGA (The Cancer Genome Atlas), as it (also) contains a large number of patients and provides survival information.

Table~\ref{tab:data} lists some of the important characteristics of these datasets. 
Note that KIPAN contains 15,529 genes, 
while METABRIC has 49,576 probes.
This is because many METABRIC probes may correspond
to the same gene each targeting a different DNA segment of the gene.
As different probes for the same gene might behave differently, 
we gave our learning algorithm the complete set of probes. 
Our results on the KIPAN dataset show that our approach also works when dealing with gene expression data from a totally different cancer
and
platform (here kidney not breast,
and mRNAseq rather than Microarray)
-- demonstrating the generality of our approach. 

\subsection{Training vs Test Data} 
\label{sec:train_val}
We apply the same experimental procedure to both datasets (METABRIC and KIPAN):
We partition each dataset into two subsets,
and use 80\% of the data for training and the remaining 20\% for testing. 
Both partitions contain instances with comparable ranges of survival times and 
comparable censored-versus-uncensored ratio.
When necessary, we ran internal cross-validation, within the training set,
to find good settings for parameters, etc. 

\section{Overview of learning and performance processes} 
\label{sec:overview_LP}

\begin{figure}[t]  
\centering
\includegraphics[scale=0.35]{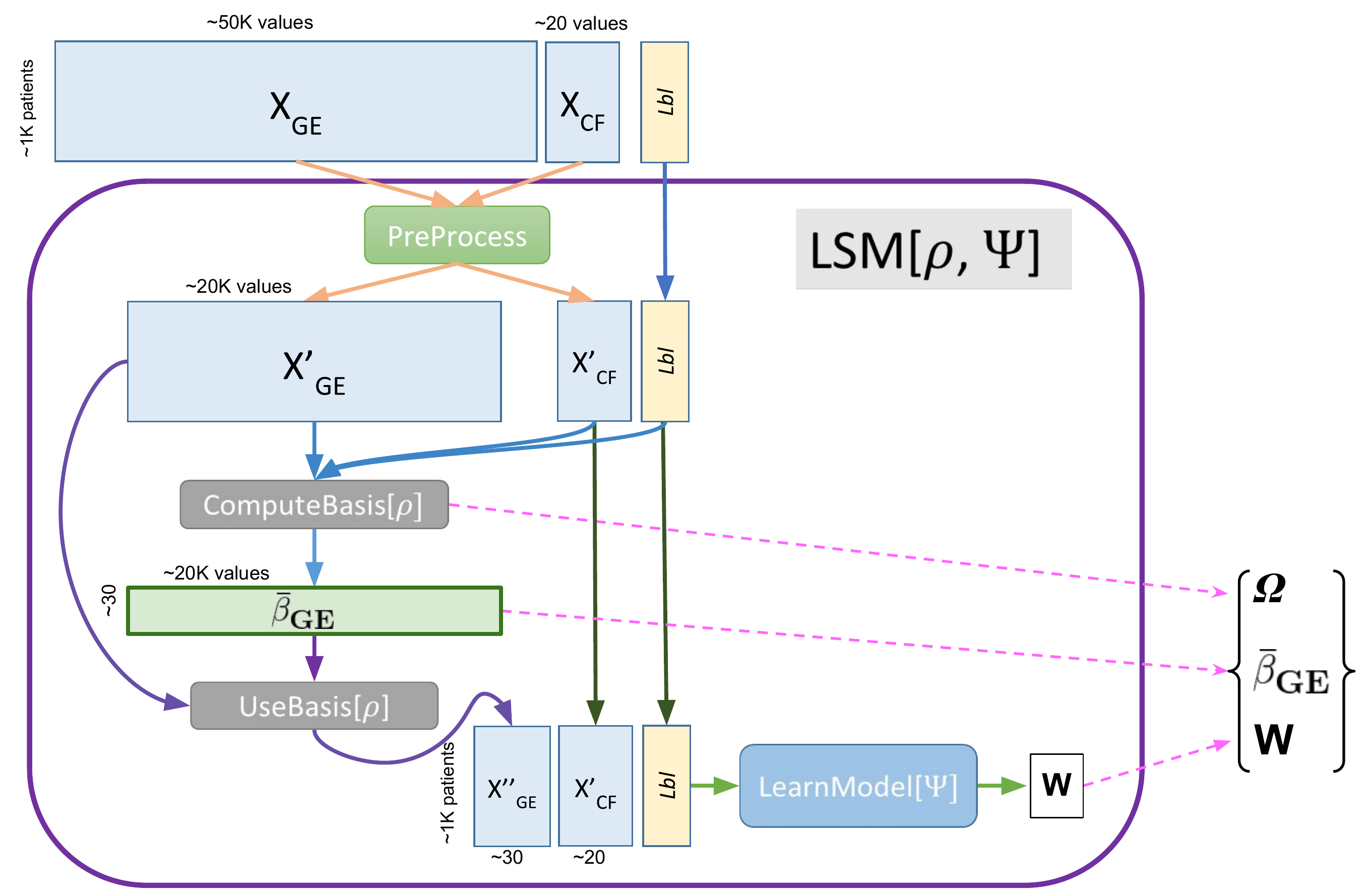}
\caption{
\label{fig:LSM}
Overview of Learning Process: LSM = LearnSurvivalModel\\
Uses $\Psi \in \{$ MTLR, Cox, RCox $\}$ for type of learner;
$\rho \in \{$ dLDA,~PCA~\} for the type of basis;
$X_{GE}$ are the gene expression values,
$X_{CF}$ are clinical features for a set of patients,
``$\Lbl$'' is the set of their (survival prediction) labels, 
$\wordProb_{GE}$ is the basis, of type $\rho$, based on the instance $X_{GE}$;
$W$ is the learned survival model, of type $\Psi$; and $\Omega$ is information about the preprocessing.
Each (unrounded) box corresponds to data, 
whose dimensions appear around it. Each row is a patient, and each column, a feature. Each rounded box is a subroutine; here we show the input and output of each.
}
\end{figure}

As typical for Supervised Machine Learning systems,
we need to define two processes:

\begin{itemize}
\item The learning algorithm, 
\add[R1]{LearnSurvivalModel}\\
\hspace*{0.1in}
\ParamAlg{LSM}{[$\rho$=dLDA; $\Psi=$MTLR]} 
{(\,[$X_{GE}$, $X_{CF}$], $\Lbl$\,) }\\
takes a labeled dataset, 
involving both gene expression data $X_{GE}$ and clinical features $X_{CF}$ 
(and survival-labels $Lbl$) for many patients,
and computes a 
$\Psi=$MTLR
survival model $W$.
\footnote{ 
  Many subroutines are parameterized by a dimensionality reduction technique $\rho \in$\{dLDA, PCA\},
  and/or by a survival learning algorithm $\Psi\in$\{MTLR, Cox, RCox\}.
We use notation
``\ParamAlg{Alg}{[$\rho$; $\Psi$]}{($\,\cdot\,$)}'' 
to identify the specific parameters;
hence \ParamAlg{LSM}
 {[$\rho$=dLDA; $\Psi=$MTLR]}  
{(\,[$X_{GE}$, $X_{CF}$], $\Lbl$\,) }
is dealing with the $\rho=$dLDA encoding and $\Psi=$MTLR survival learning algorithm.
}
It also returns the 
$\rho=$dLDA
``basis set'' $\wordProb_{GE}$ 
(here, think of a set of \strain{} distributions),
and some information about the pre-processing performed, $\Omega$.
See Figure~\ref{fig:LSM}

\item The performance algorithm, \add[R1]{UseSurvivalModel}\\
\hspace*{0.1in}  \ParamAlg{USM}
{[$\rho$=dLDA; $\Psi=$MTLR]} 
{( [$x_{GE},\ x_{CF}],\ \wordProb_{GE},\ W,\ \Omega$~)}, \\
takes a description of an individual (both gene expression $x_{GE}$, and clinical features $x_{CF}$), 
as well as the
$\rho=$dLDA
basis set $\wordProb_{GE}$ 
and the 
$\Psi=$MTLR survival model $W$
(and pre-processing information $\Omega$), and returns a specific survival prediction for this individual, from which we can compute that person's risk score.
See Figure~\ref{fig:usm_workflow}. 
\end{itemize}
To simplify the presentation, 
the main text will describe the process at a high-level,
skipping most of the details.
Notice these functions are parameterized by the 
type of dimensionality reduction $\rho$ and 
the survival learner $\Psi$.
This section will especially focus on the novel aspects here,
which are
the $\rho\,= \,$dLDA transformation (Section~\ref{sec:dLDA}),
which complicates the 
\ParamAlg{\ComputeBasis}  {[$\rho\,=\,$dLDA]}{($\cdots$)}\
function
(Section~\ref{sec:ComputeBasis-dLDA});
and the $\Psi\,=\,$MTLR algorithm for learning the survival model
(Section~\ref{sec:MTLR}).
Appendix~\ref{app:Details-Alg} summarizes the more standard 
$\rho\,=\,$PCA approach to reducing the number of features,
 and the more standard  survival models $\Psi \in\,$\{Cox, RCox\},
 as well as other details about the learning, and performance models, in general.

\subsection{Learning System \ParamAlg{LSM}{}{}} 
\label{sec:LSM}

Here, {\tt LSM{\scriptsize [$\rho =$ dLDA; $\Psi=$  MTLR]}( [$X_{GE}$, $X_{CF}$], $\Lbl$~)} 
first calls \PreProcess{},
which fills-in the missing values in the $X_{CF}$ clinical features (producing $X'_{CF}$),
and normalizes the real-valued $X_{GE}$ genetic features,
which is basically computing the z-scores
$X'_{GE}$,
over {\em all}\ of the values.
It then calls 
\ParamAlg{ComputeBasis}{[$\rho\ =$ dLDA]}{($\cdots$)}\
to compute a set of  \strain{}s  $\wordProb_{GE}$
from the gene expression data $X'_{GE}$
(as well as the other inputs),
then calls 
\ParamAlg{UseBasis}{[$\rho\ =$ dLDA]}{(~$X'_{GE},\ \wordProb_{GE}$~)}, 
which ``projects'' $X'_{GE}$ onto this $\wordProb_{GE}$ to find a low dimensional description of the genetic information;
see Figure~\ref{fig:C-U-Basis}.
These projected values, together with $X'_{CF}$ and  $\Lbl$,
form the labeled training set
given to the $\Psi=$MTLR learning system, 
which computes a survival model $W$.
Here, the {\tt LSM} process returns the 
dLDA ``basis'' $\wordProb_{GE}$ 
and the MTLR-model $W$.
(Further details appear in Appendix~\ref{app:Details-Alg}.)

\subsubsection{\ParamAlg{ComputeBasis}{[$\rho\ =$ dLDA]}{($\cdots$)}
function} 
\label{sec:ComputeBasis-dLDA}
As noted above, the $\approx$50,000
expression values for each patient 
is so large that most standard learning algorithms would overfit.
We consider two ways to reduce the dimensionality.
One standard approach,
Principal Component Analysis (PCA),
is discussed in Appendix~\ref{sec:superpc+}.
Here, we discuss a different approach,
dLDA,
that uses the Latent Dirichlet Analysis.

The \PreProcess\ routine computes z-scores $X'_{GE}$ for the gene expression values $X_{GE}$;
the \ParamAlg{\ComputeBasis}{[$\rho\,=\,$dLDA]}{}\ subroutine then has to 
transform those real values to the non-negative integers required by LDA
--
moreover, it was designed to deal with word counts in documents where, 
in any given document, 
most words appear 0 times, then many fewer words appear once, then yet fewer words appear twice, etc.
We therefore need a method for converting the 
  real values 
into non-negative integers.

 \begin{figure}\centering
 \includegraphics[width =\textwidth,height=2.5in]{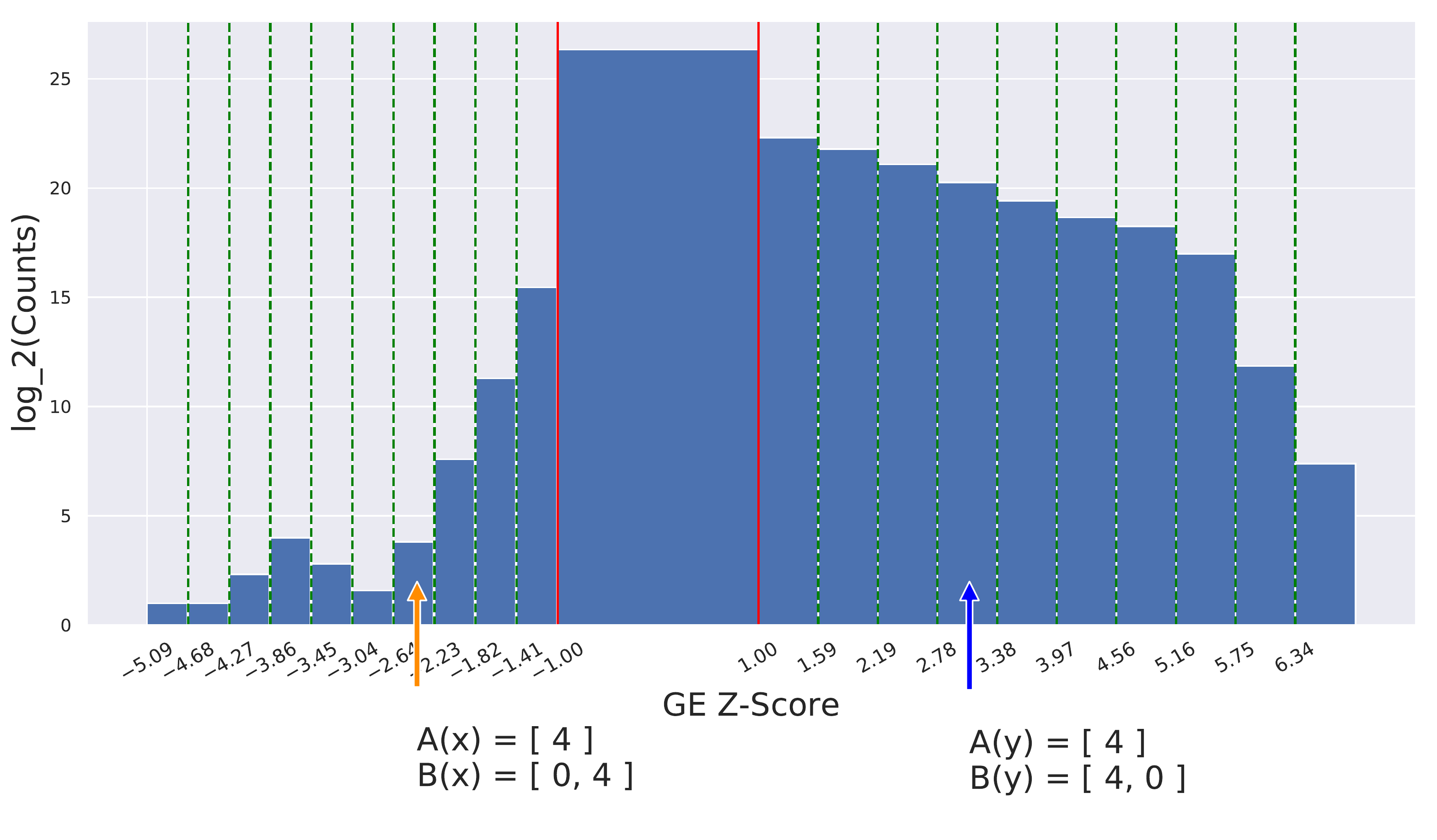}
\caption{
\label{fig:binning}
Histogram of the normalized Gene Expression values
$\{x_i^j\}$ , from METABRIC,
showing how we descretized them
into essentially equal-width bins.
Note the heights are on a log-scale.\\
The material under the histogram
-- involving {\bf x} and {\bf y} --
compare two ways to compute the discretized Gene Expression Values (dGEV):
The top \tech{A}\ discretizes the GEVs into a 
single count-feature A($\cdot$),
and the bottom \tech{B}\ discretizes the GEVs into two count-features B($\cdot$) representing over-expression and under-expression, respectively.}
\label{fig:encoding_AB}
\end{figure}

This process therefore {\em discretizes}\
the standardized gene expression values (in $X'_{GE}$)
into 
the integers
\{ -10, -9, \dots, -1, 0, 1, \dots, 9, 10 \},
by  mapping each real number to the integer indexing
some essentially equal-sized bins;
see Figure~\ref{fig:binning},
and Appendix~\ref{sec:ComputeBasis} for details.

This does map each gene expression to an integer, but this 
includes both positive and negative values.
Given that over-expression is different from under-expression,
an obvious encoding uses two non-negative integer values for each gene: 
mapping $+$2 to [2\,,0], and $-$3 to [0,\,3], etc.
Note that the range of each component of the encoding will be non-negative integers,
and that most of the values will be 0, then fewer will be 1, etc.\ -- as desired.
{However, this does double the dimensionality of representation;
}
\add[R1]{
\hbox{\ie} we now have twice the number of genes:
 UNDER-`gene\_name' and OVER-`gene\_name'.
}
(Below we call this
the \tech{B}\ encoding; 
see also $B(\cdot)$ at the bottom of
Figure~\ref{fig:encoding_AB}.)
Given that very few values are $<\ -1$
 (in METABRIC, over 14\% (normalized) expression values were $>1$,
  but less than less than 0.04\% were $<\ -1$;
 recall that heights in Figure~\ref{fig:encoding_AB}
 are on a log scale), we considered another option: 
collapsing the $+$values and $-$values to a single value --
so both $+$4 and $-$4 would be encoded as 4.
This would mean only half as many features 
(which would reduce the chance of overfitting),
and would continue to note when a gene had an exceptional value.
(This is 
the \tech{A} encoding,
which corresponds to the $A(\cdot)$ at the bottom of
Figure~\ref{fig:encoding_AB}.)
As it was not clear which approach would work better,
our implementation explicitly considered both options --
and used the training set to decide which worked best; 
see below.

The standard LDA algorithm also 
needs to know the number of topics (here \strain{}s) $K$ to produce.
\ComputeBasis{} uses (internal) cross-validation to find the best value for $K$, over the range $K \in \{5,\ 10,\ 15, \dots, 150\}$,
as well as encoding technique $t \in \{$\tech{A}, \tech{B}$\}$
-- seeking the setting leading to the Cox model with the best concordance
(on each held-out portion).
See Appendix~\ref{sec:ComputeBasis} for details.
After finding the best $K^*$  and encoding $t^*$,
\ComputeBasis\ then finds the $K^*$ \strain{}s
on the $t^*$-encoded 
(preprocessed) 
training gene expression data $X'_{GE}$; 
this is the \strain{} distribution, $\wordProb_{GE}$.

\begin{figure}[t]\centering
 \includegraphics[width =5in, height=2.2in]{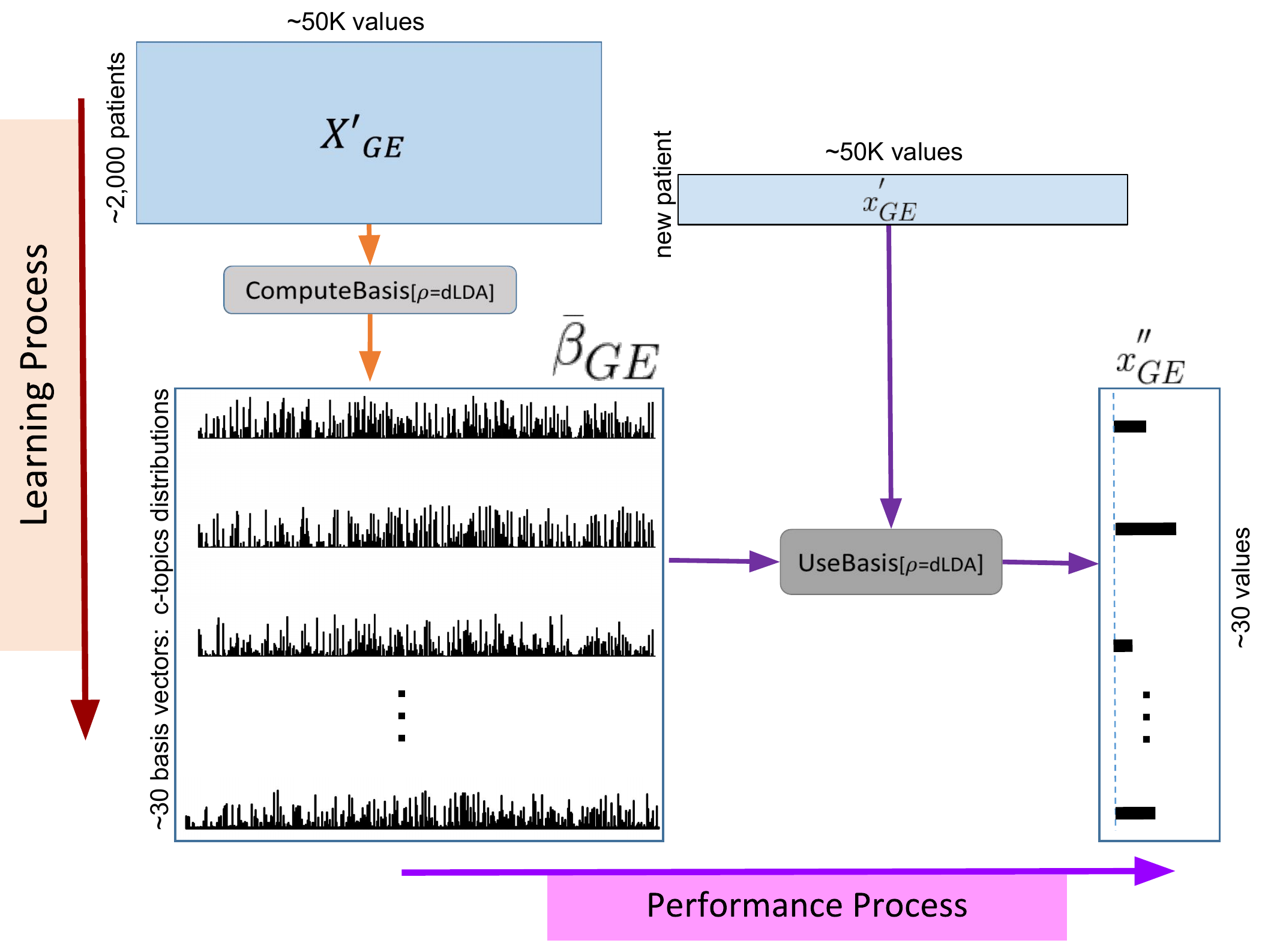}
\caption{
\label{fig:C-U-Basis}
The \ParamAlg{ComputeBasis}{[$\rho=$dLDA]}{}\ 
process 
(shown top-to-bottom, on left)
uses a set of 
high-dimensional
gene expression vectors
$X'_{GE}$ from many patients,
to produce a set of basis vectors
(corresponding the parameters of the LDA) $\wordProb_{GE}$.
The \ParamAlg{UseBasis}{[$\rho=$dLDA]}{}\ 
process
(shown left-to-right horizontally)
uses the set of dLDA ``basis vectors''
$\wordProb_{GE}$
to transform a gene expression vector $x'_{GE}$
from a novel patient,
into a low-dimensional description $x''_{GE}$.
(Note \ParamAlg{ComputeBasis}{[$\rho=$dLDA]}{}\ 
also uses other information,
$X'_{CF}$ and $Lbl$, to determine the number $K$
of basis vectors; 
to simplify the figure, we did not show this.)
}
\end{figure}

The vertical left-side of Figure~\ref{fig:C-U-Basis}
gives a high-level description of the 
\ParamAlg{ComputeBasis}{[$\rho=$dLDA]}{}\ 
process:
given a large set of
(preprocessed) high-dimensional gene expression profiles,
produce a small set of \strain{}s
(each corresponding to a mapping from the gene expression profiles).
We will later describe the 
\ParamAlg{UseBasis}{[$\rho=$dLDA]}{}\ 
process that uses those \strain{}s to 
transform the high-dimensional gene expression profile of a novel instance,
into a small dimensional set of values
-- see the left-to-right 
``Performance \add[m]{Process}'' part here.
At this abstract level, 
it is easy to see that it nicely matches the
$\rho=$PCA process,
where \ParamAlg{ComputeBasis}{[$\rho=$PCA]}{}\ 
would find the top principle components of the $X'_{GE}$ datasets
(here, the $\wordProb_{GE}$ box would be those components),
which  \ParamAlg{UseBasis}{[$\rho=$PCA]}{}\ 
could then use to transform a new gene expression profile into that low-dimensional ``PC-space''.

\subsection{Performance system, {\tt USM}} 
\label{sec:USM}

\begin{figure}[t] 
\centering
\includegraphics[width=0.8\textwidth,height=2.5in]
{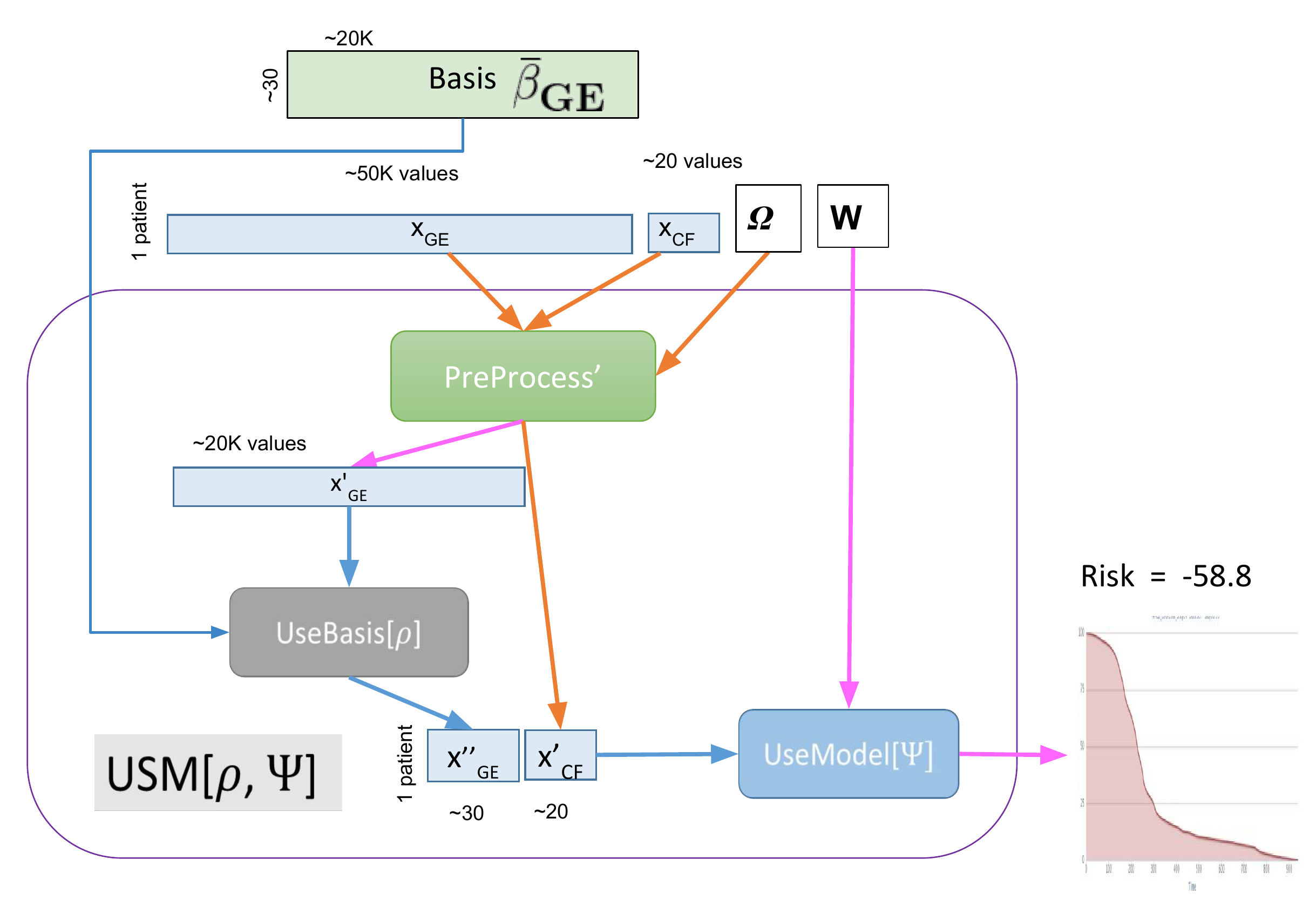}
\caption{
Overview of Performance Process, USM = {\tt UseSurvivalModel}.
$x_{GE}$ and $x_{CF}$ are the gene expression values and clinical feature values, for a single patient; see also terms $\rho$, $\Psi$, $\Omega$, $\wordProb_{GE}$, $W$ from Figure~\ref{fig:LSM}.
\label{fig:usm_workflow}
}
\end{figure}

As shown in Figure~\ref{fig:usm_workflow},
the 
\ParamAlg{USM}{[$\rho=$dLDA;$\Psi=$MTLR]}{(~$[x_{GE},\ x_{CF}],\ \wordProb_{GE},\ W,\ \Omega$~)}\
system applies the learned $\Psi=$MTLR model $W$,
to a \PreProcess{}'ed description of a novel patient,
$[x''_{GE}, x'_{CF}]$,
whose gene expression values $x''_{GE}$
have been 
``projected'' into the relevant basis $\wordProb_{GE}$
by
\ParamAlg{UseBasis}{[$\rho=$dLDA]}{}.
This produces a survival curve,
which 
it then
uses
to produce that patient's predicted risk score:
the negative of the expected time for this distribution,
which corresponds to the area under its survival curve.

Each of the various subroutines are described in an appendix:
\ParamAlg{\PreProcess{}'}{}{}, 
\ParamAlg{\UseBasis}{[$\rho=$dLDA]}{}\ and \ParamAlg{\UseBasis}{[$\rho=$PCA]}{}\
are described in Appendices~\ref{app:PreProcess}, \ref{sec:UseBasis} and \ref{sec:superpc+}, respectively.
The
\ParamAlg{\UseModel}{[$\Psi=$Cox]}{}\ and \ParamAlg{\UseModel}{[$\Psi=$RCox]}{}\
produce standard risk scores for each patient \change[m]{ using}{, obtained by applying 
the learned Cox (resp., RCox) model to}
 the patient's clinical and gene expression features;
see Appendix~\ref{sup:cox}. 

\section{Experimental Results} 
\label{sec:results}
As noted above,
we intentionally designed our learning and performance systems 
(Figures~\ref{fig:LSM} and \ref{fig:usm_workflow})
to be very general
-- to allow two types of basis
 $\rho \in \{\,$dLDA, PCA$\,\}$
 and three different survival prediction algorithms 
$\Psi \in \{\,$MTLR, Cox, RCox$\,\}$.
This allows us to explore 
$2 \times 3$ frameworks,
on the two different datasets (METABRIC and KIPAN).
For each, the learner uses internal internal cross-validation to find the optimal parameters.
Below we report the results of each optimized model
on the held-out set,
focusing on the Concordance Index \add[R1]{(CI)}
--
a discriminator measure.
We also discuss a {\em calibration}\ measure of these results;
see Appendix~\ref{app:d-calib}.

We also present our experimental results from the BCC Dream Challenge winner's model~\cite{cheng2013development}.
As discussed in Section~\ref{sec:ComputeBasis-dLDA},
\ParamAlg{ComputeBasis}{[$\rho=$dLDA]}{}\ ran internal cross-validation on the training set
to determine the appropriate encoding $t^* \in \{$\tech{A}, \tech{B}$\}$
and  
the optimal number of \strain{}s for the dLDA model $K^*$
from a large potential values (see Algorithm~\ref{alg:dLDA} in 
Appendix~\ref{sec:ComputeBasis}).
Our experiments found that the discretization $t^* =$ \tech{B}, 
along with $K^*\! =\, $30 \strain{}s, produced the best dLDA algorithm for survival prediction in METABRIC; 
after fixing the encoding scheme as \tech{B}, we used the same technique on the KIPAN dataset and found $K\!=\,$50 \strain{}s to be the best.
\add[R1]{We used the 
C implementation from Blei~\hbox{\etal~\cite{blei2003latent}}%
\footnote{https://github.com/blei-lab/lda-c}
to compute the \strain{}s.
On a single 2.66GHz processor with on 16Gb memory,
a single fold takes around $\sim$20 -- 30 hours (more time for larger $K$). We, of course, parallelized each CV fold.}

We experimented with different combinations of the features from three groups:
(1)~clinical features,
(2)~SuperPC+ principle components ($\rho=$PCA), 
and/or 
(3)~the dLDA \strain{} ($\rho=$dLDA); 
and with three different survival prediction algorithms 
$\Psi\in \{$ Cox, Cox, MTLR $\}$.
Our goal in these experiments is to empirically 
evaluate the performance of the survival models that use various types of features.
Given this goal, we evaluate the performance using different GE basis methods ($\rho$)
by comparing their performance to a baseline model that only uses the clinical features with Cox~\cite{coxmodel}.
The other combinations include clinical features as well as various different GE features; 
each is trained using each of the three aforementioned survival prediction algorithms ($\Psi$).

\def\sCox{\hspace*{-1em}Cox\hspace*{4em}}
\def\sRCox{\hspace*{0em}
  RCox\hspace*{1em}}
\def\sMTLR{\hspace*{3em} MTLR}

\def\NA{NA} 

\begin{table*}
\begin{adjustwidth}{-1.0in}{0in} 
\centering 
\caption{Concordance results of various models from METABRIC 
(over 395 test instances) and KIPAN 
(over 176 test instances). 
\\
\pp\ = used these features and 
\mm\ = did not use these features\\
As PAM50 is specific to breast cancer,
it is {\it not applicable}\ to the kidney (KIPAN) data.\\
The first row, with the ID ``A(*)'', is the baseline.
} 
\label{tab:metabric}
\small
\begin{tabular}{@{}l|cccc|c|cc} 
\hline 
\rowcolor{gray!20}
ID  & \multicolumn{4}{c|}{Feature Groups} 
& Learning Alg. 
& \multicolumn{2}{c}{Concordance} \\ 
\rowcolor{gray!20}
 & Clinical & PCA & dLDA & PAM50 & 
 {\tiny Cox $|$ RCox $|$ MTLR} & {\scriptsize METABRIC} & {\scriptsize KIPAN} \\ 
\thickhline
A (*) 
& \pp & \mm & \mm & \mm & \sCox & 0.6810 & 0.7656  \\ 
A  & \pp & \mm & \mm & \mm & \sRCox\qquad & 0.6883 & 0.8156 \\ 
A  & \pp & \mm & \mm & \mm & \sMTLR & 0.6820 & 0.8207 \\ 
\hline
B  & \pp & \pp & \mm & \mm & \sCox & 0.6961 & 0.7691 \\ 
B  & \pp & \pp & \mm & \mm & \sRCox & 0.7048 & 0.8196 \\ 
B  & \pp & \pp & \mm & \mm & \sMTLR & 0.6999 & 0.8232 \\ 
\hline
C  & \pp & \mm & \pp & \mm & \sCox & 0.7073  & 0.7638 \\
C  & \pp & \mm & \pp & \mm & \sRCox & 0.7108 & 0.8332  \\ 
C  & \pp & \mm & \pp & \mm & \sMTLR & 0.7139 & 0.8482 \\ 
\hline
D  & \pp & \pp & \pp & \mm & \sCox & 0.7074 & 0.8062 \\
D  & \pp & \pp & \pp & \mm & \sRCox & 0.7145 & 0.8401 \\ 
D  & \pp & \pp & \pp & \mm & \sMTLR & 0.7079 & {\bf 0.8495} \\ 
\hline
E  & \pp & \pp & \pp & \pp & \sCox & 0.7075  & \NA  \\
E  & \pp & \pp & \pp & \pp & \sRCox & 0.7172  & \NA \\
E  & \pp & \pp & \pp & \pp & \sMTLR &\ {\bf 0.7202} & \NA \\ 
\hline
F  & \pp & \multicolumn{3}{c|}{Meta-Genes} & Ensemble Model & 0.7293 & \NA\\
\hline 
\end{tabular}
\end{adjustwidth}
\end{table*}

\begin{figure*}[t]\small

\begin{tabular}{ll}
\includegraphics[width=2.5in, height=1.5in]{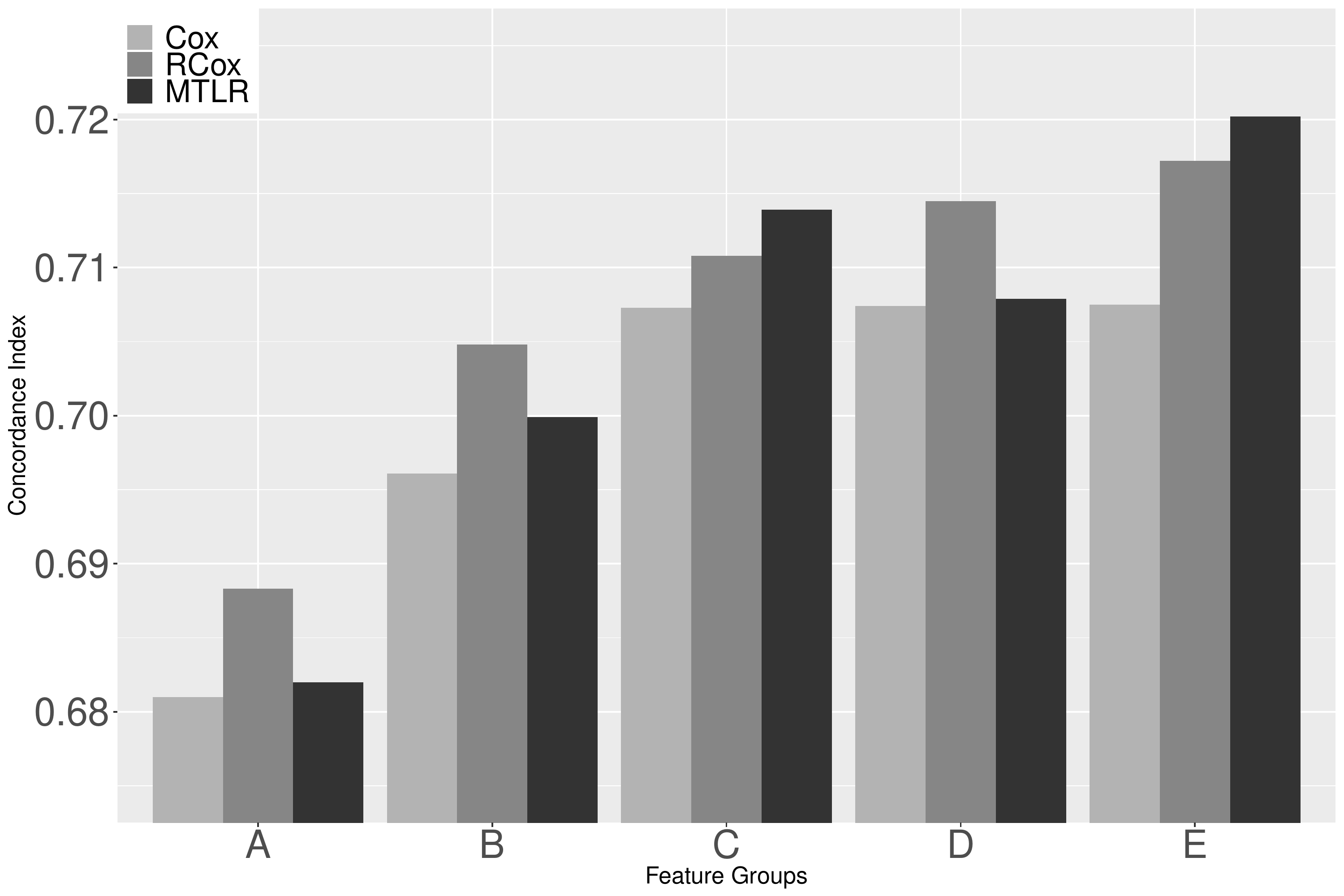}
&
\includegraphics[width=2.5in, height=1.5in]{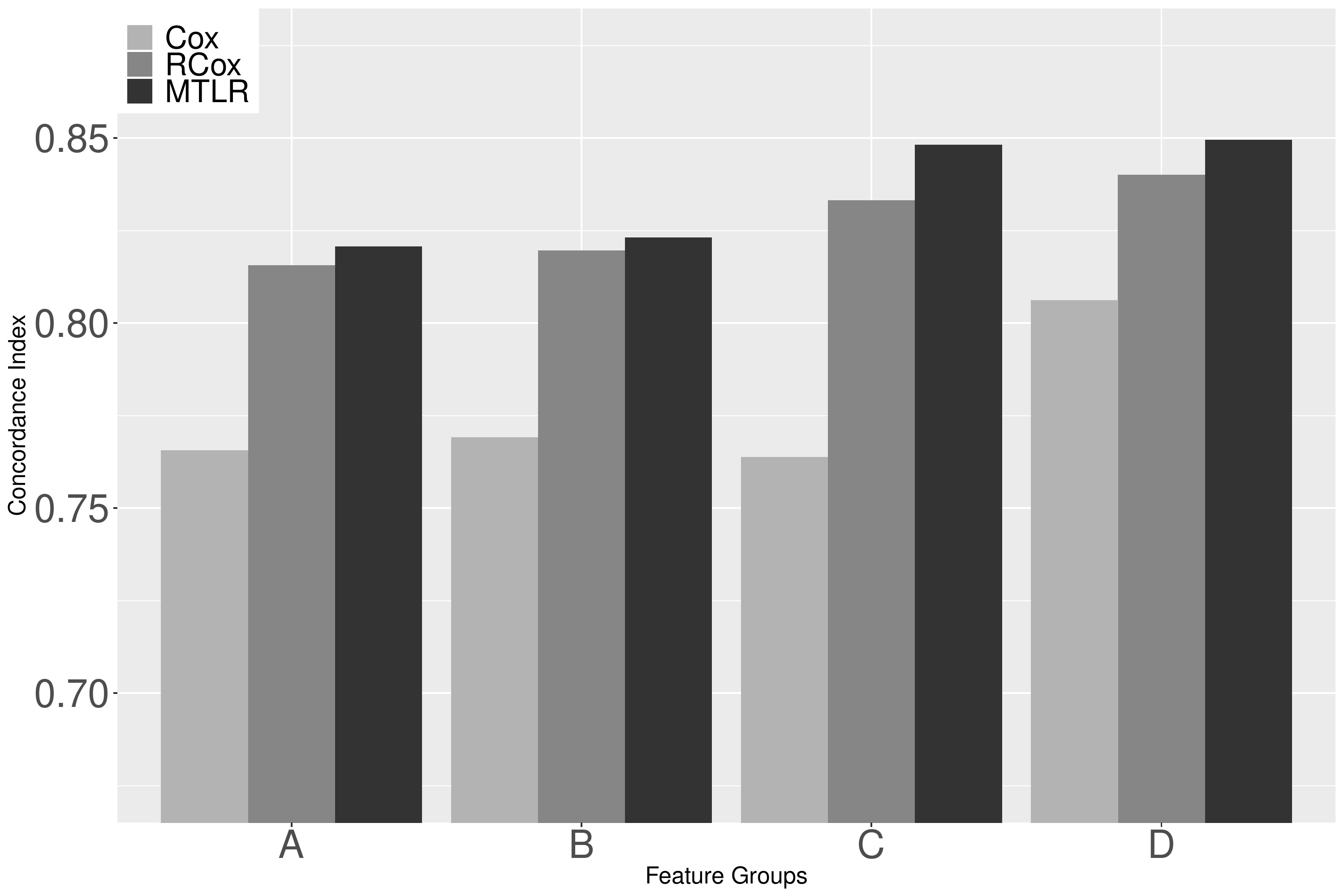}
\end{tabular}
\caption{Test CI: METABRIC (left) and KIPAN (right). 
Note higher values are better.
The labels on the x-axis correspond to
Table~\ref{tab:metabric}.
}
\label{fig:barplot}

\end{figure*}

As an additional feature selection step, 
we removed the covariate ``Site'' from the METABRIC clinical covariates, 
based on 
our
experimental results (on the training data)
that shows its inclusion led to worse concordance. 
We experimented with a large, but selective,
set of model combinations,

to answer our major queries:\\[-1em]

\begin{enumerate}
\item[(i)] does adding GE features improve survival prediction?

\item[(ii)] which is the best feature combination for
 
survival prediction?

\item[(iii)] which is better representation of the GE features: dLDA or SuperPC+?

\item[(iv)] are we deriving GE features that are redundant with PAM50?

\end{enumerate}

Our results appear in Table~\ref{tab:metabric},
shown visually in Figure~\ref{fig:barplot}(left).
These lead us to claim:\\
(i)~Comparing the baseline, A-Cox%
\footnote{To explain this notation,
the `A' refers to the feature set used,
which here is the far left triplet of blocks in Figure~\ref{fig:barplot}(left); 
the `Cox' refers to the learning algorithm,
which appears left-most in each triplet.
See also Table~\ref{tab:metabric}.
},
to the other models, 
we immediately see that adding GE features
(using any of the dimensionality reduction technique) 
leads to better predictive models;
-- \ie all of the results are better than 
A-Cox's CI of 0.6810  
(the left-most light-shaded bar in Figure~\ref{fig:barplot}(left) ).\\
{(ii)~The best model for METABRIC is the one that includes all of the types of features derived from the gene expression -- here E-MTLR,
which is the right-most bar of Figure~\ref{fig:barplot}(left).

We also performed student's t-tests on random bootstrap samples from the test data to validate the significance of our results.
When we compare this best model, E-MTLR, 
against models B-RCox (which is the best model using only PCA GE features) and C-MTLR (the best model using only dLDA GE features),
we find statistically significant difference between them (respective pairwise p-value: 4.8e-16, 1e-3), 
showing that the  E-MTLR model
is significantly better than its closest counterparts.\\
(iii)~These empirical results show that, 
if you are pick only a single GE feature set, 
the dLDA \strain{}s perform better than the principle components --
that is, the C-$\chi$ has a higher score than B-$\chi$,
for $\chi \in \{$Cox, RCox, MTLR$\}$; 
moreover, a model using both sets of features performs yet better
(\ie D-$\chi$ is better that C-$\chi$).
\\
(iv)~Comparing the D-$\chi$ to E-$\chi$,
we see that adding PAM50 subtypes as features to the 
METABRIC database improves  the held-out test concordance. 

Indeed, we see that the performance of models that include PAM50 are marginally
better than similar models that do not (row D), suggesting that the information added by these 
different representations of GE data are not redundant.
Moreover, 
we see that, in all feature groups, both RCox and MTLR clearly outperform Cox --
\ie $\nu$-RCox and $\nu$-MTLR are  better than $\nu$-Cox,
for $\nu \in \{$A,B,C,D,E$\}$.
We then tested the first three claims on the KIPAN dataset;
see Table~\ref{tab:metabric} (right-most column) and Figure~\ref{fig:barplot}(right).
(As KIPAN does not deal with breast cancer,
the PAM50 features are not relevant,
so we could not test claim~(iv).)\\
(i)~As before, we found that adding expression information improves over the baseline A-Cox --
\ie essentially all values are better than 0.7656.
\\
(ii)~We again found that the best model was the one that included all of the features; here D-MTLR. 
Moreover, a t-test on bootstrap replicas show that this model D-MTLR was {\em significantly}\ better than the top model that does not include dLDA features, B-MTLR.
\\
(iii)~We again see that C-$\chi$ has a higher score than B-$\chi$, meaning (again) that models trained with 
\emph{only the \strain{}s performed much better than PCA-features}; 
but that including both features was yet better (D-$\chi$).

These sets of experiments support our claim that
\begin{eqnarray*}
&&\hbox{a model learned by running MTLR on all GE features,}\\[-0.5em]
&&\hbox{gives very good concordance scores}   
\end{eqnarray*}
-- statistically better than other options in two
different datasets, using different platforms,
related to different cancer types. 

In addition to these evaluations using the discriminative concordance measure,
we also applied a calibration measure:
``D-Calibration''
(``D'' for ``Distribution'')~\cite{AndresPLoSONE18,ISD-Paper},
which measures how well a individual survival distribution model is calibrated, 
using the Hosmer-Lemeshow~(HL)~\cite{hosmer2013applied} 
goodness-of-fit test;
see Appendix~\ref{app:d-calib}.
We found that all of our models,
for both datasets (METABRIC and KIPAN),
passed this calibration test; 
see Appendix~\ref{sup:calibration}, especially Table~\ref{tab:calib}. 
But we have found that this is not universal.
For example, we experimented with another breast cancer dataset BRCA
(results not shown here), 
and found that the Cox model failed for all configurations (of $\rho$), 
showing that the Cox model does not always produce calibrated results -- 
here, for situations where RCox and MTLR produced D-calibrated predictors.
See also Haider~\etal~\cite{ISD-Paper}.
 
\subsection{Other Comparisons} 
\label{sec:OtherComparisons}

In 2012, Cheng~\etal~\cite{cheng2013development} won the BCC Dream Challenge 
(which was based  on the METABRIC data) by  (i)~leveraging prior knowledge of cancer biology to form Meta-Genes and (ii)~training an ensemble of multiple learners, fueled by the continuous insights from the challenge competitors via open sharing of code and trained models. 
To compare our performances with this BCC  winning program,
we reproduced their models (using the DreamBox7 package),
then re-trained their ensemble learners on our training split of the METABRIC data
and tested on the held-out test set.
Table~\ref{tab:metabric}[Row~{\em F}]
shows that the resulting ensemble model achieved a CI of 0.7293 on the test data.
While that score is slightly better than the performance of our best model (Table~\ref{tab:metabric}[Row~{\em E}]),
note that all of our tuning was performed solely on the training (n=1586) data,
while their team 
made major design choices for their model
using the entire METABRIC cohort (all n=1981 instances),
on which it was then evaluated.

Recently, Yousefi~\etal~\cite{yousefi2017predicting} 
trained a deep neural network 
on this KIPAN data -- 
including this gene expression data, as well as
other features: Mutation, CNV and Protein.
They reported concordance scores around 0.73 $-$ 0.79,
which are 
lower than our best, 0.8495 .

\add[R2]{Finally, while we focused on the LDA approach,
we also explored another topic modeling technique,
Latent Semantic Indexing (LSI)~\hbox{\cite{hofmann2001unsupervised}}.
Running this on both datasets
(using the same discretization approach,
the same $t^* =\,$\hbox{\tech{B}} encoding and
the same number of \hbox{\strain}s, $K^* =\,$30),
we found essentially the same Concordance values,
and confirmed that all four claims
(i) through (iv) still hold, just replacing dLDA with 
the ``discretized LSI'' (dLSI) encoding.
}

\section{Discussion} 
\label{sec:discussion}
Given the growing number of gene expression datasets 
as part of survival analysis studies,
it is clearly important to develop survival prediction models that can utilize such high-dimensional GE data. 
This motivated us to propose a novel survival prediction methodology 
that can learn predictive features from such
GE data -- exploring ways to learn and use \strain{}s
as features for models that can effectively predict survival.
{\em  N.b.}, this paper focuses exclusively on this predictive task,
 as this can lead to clinically relevant patient-specific information;
 indeed, this motivated the BCC Dream challenge, which provided the METABRIC dataset.
We anticipate future work will explore the possible interpretation of these \strain{}s.

We included Cox as one of our learning modules for this 
task as it is known to be effective
at optimizing concordance, both empirically and theoretically~\cite{steck2008ranking}.
We included RCox as this algorithm recently won Prostate Cancer Dream Challenge 9.5~\cite{guinney2016prediction}.
Finally, we included the MTLR survival prediction model
as its performance, there, was competitive with the best,
as well as based on the empirical evidence in 
Haider~\etal~\cite{ISD-Paper}.
Our evaluations on these two datasets show that 
\change[m]{its}{MTLR's} 
performance was often better than RCox and Cox.
Moreover, 
while the basic RCox and Cox functions produce only a risk score for each patient,
MTLR
provides
a survival distribution for each,
mapping each time to a probability;
see Figure~\ref{fig:pssp_surv}.
Such models, which produce an individual survival distribution, can be used 
to compute a risk score, allowing them to be used for concordance-tasks;
they can also be used to predict single time probabilities
(\eg probability of a patient living at least 3 years),
and also can be visualized.\footnote{
We did use the Kalbfleisch-Prentice approach to estimate the base hazard function,
allowing Cox and RCox to similarly produce individual survival curves.
Appendix~\ref{app:d-calib} describes a way to evaluate such ``individual survival distribution'' models,
D-Calibration.
Appendix~\ref{sup:calibration} then
shows that, for these datasets, 
these models all pass this test.
}

\remove[R3]{This paper shows that this MTLR model is a novel addition to the suite of survival prediction techniques,
that works well in general.
More information can be found in~\hbox{\cite{Yu_al_NIPS11,ISD-Paper}}.
}
\note[R3]{The following material is new:}

To summarize the main disadvantages and advantages of our approach,
versus more standard approaches (\eg PCA for dimensionality reduction, and (R)Cox for survival prediction):

\begin{itemize}
 \item Disadvantages:
   \begin{itemize}
      \item Topic models are not simple to describe.
      \item This approach requires a fairly long
        training time ($\sim$20 hours on a 16GB, 2.66GHz processor
for a single model) -- to first find the parameters (encoding, number of \strain{}s), then the \strain{}s themselves, and finally, to learn the model that has the best performance.
(However, using the trained dLDA model to predict \strain{} contributions for a new patient is very fast -- under a second on a general purpose laptop computer.)
    \end{itemize}
    
    \item Advantages:
    \begin{itemize}
        \item 
        An effective process to learn representation from gene expression data, as a meaningful probability distribution over the genes. 
        \item The learned representation from the gene expression data improves survival prediction, over standard methods, in:
        \begin{itemize}
        \item Different cancer types: Breast and Kidney.
            \item Different gene expression data types: Microarray and mRNASeq.
            \item Different survival prediction algorithms: Cox, 
            Regularized-Cox and MTLR.
        \end{itemize}
        \item Our combined approach for feature learning and survival prediction (dLDA + MTLR) archives strong concordance scores compared to standard survival models across different cancer types.
    \end{itemize}

\end{itemize}

\section{Conclusion} 
\label{sec:conclusion}
 
Table~\ref{tab:metabric} shows that our proposed model, which uses MTLR to learn a model involving various types of derived GE features (dLDA \strain{}s and/or SuperPC+),
has the best concordance, in two datasets representing different types of cancer,
and two different gene expression platforms (micro-array and mRNAseq).
That table shows that adding GE features improves survival prediction and that including both dLDA \strain{}s and SuperPC+ principle components gives the most improvements across held-out datasets
\change[m]{ -- showing}{. We also found}
that the ``framework'' that produced the best model in METABRIC, was also the best in the Pan-kidney KIPAN dataset, which shows the robustness of our proposed prediction framework.
Moreover the \strain{}s extracted by our dLDA procedure 
(inspired by topic modeling) can be interpreted as
collections of over-expressed or under-expressed gene sets; 
further analysis is needed to discover and validate the biological insights from these \strain{}s. Our results show that our novel survival prediction model
-- learning a MTLR survival model based on our derived GE features
(dLDA \strain{}s and SuperPC+ components)
-- leads to survival prediction models that can
be better than standard survival models.
\add[R2]{We anticipate that others will find
this dLDA+MTLR approach
helpful for their future tasks.}





\appendix


\section{Details about the Algorithms} 
\label{app:Details-Alg}
\begin{figure}[t] 
\centering
\includegraphics[width=0.8\textwidth]{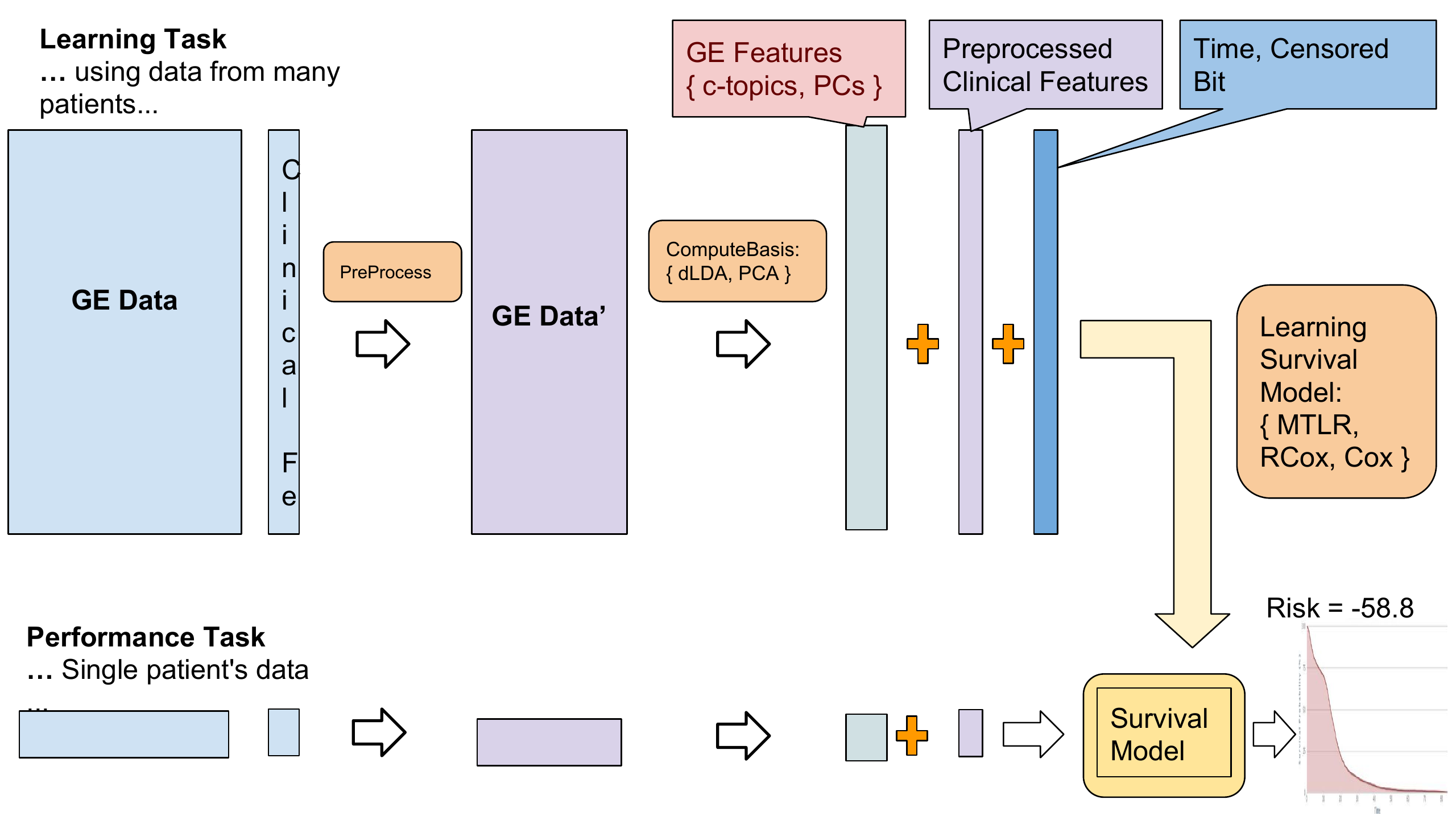}
\caption{ \label{fig:usm_workflow-both}
Simplified~flow~diagram~describing the overall prediction process, 
including both the learning task and the performance task.
}
\end{figure}
Section~\ref{sec:overview_LP} gave a high-level overview
of the important parts of the learning, and performance, systems;
see also Figure~\ref{fig:usm_workflow-both}.
This appendix completes that description.
In particular, it summarizes the components of the learning and performance systems --
each shown as a rounded-rectangle in Figures~\ref{fig:LSM}
or \ref{fig:usm_workflow}
-- roughly in a top-to-bottom fashion.
Appendix~\ref{app:PreProcess} describes the
\ParamAlg{PreProcess}{}{($\cdots$)}\ 
routine that preprocesses the training data
(both gene expression and clinical features),
and the related 
\ParamAlg{PreProcess'}{}{($\cdots$)}\ routine,
used by \ParamAlg{USM}{}{}, to preprocess a novel instance.
Appendix~\ref{sec:ComputeBasis} then gives
many details about 
\ParamAlg{ComputeBasis}{[$\rho=$dLDA]}{($\,\cdots\,$)}\
that computes the set of ``\strain\,$-$\,genes'' distributions,
given gene expression values (and some additional information)
-- extending the high-level description in
routine in Section~\ref{sec:ComputeBasis-dLDA}.
Appendix~\ref{sec:UseBasis}
describes the 
\ParamAlg{UseBasis}{[$\rho$=dLDA]}{}\ routine that uses these \strain-genes distributions to map each patient's gene expression profile into that patient's specific 
``\strain\,-\,distribution'';
see Figure~\ref{fig:C-U-Basis}.
Appendix~\ref{sec:superpc+} presents \ParamAlg{ComputeBasis}{[$\rho=$PCA]}{} 
and
\ParamAlg{UseBasis}{[$\rho=$PCA]}{} 
techniques,
to deal with the other approaches for reducing the dimensionality, PCA.
Finally Appendix~\ref{sup:cox} describes 
two related standard survival analysis methods:
Cox~\cite{coxmodel}, and Ridge-Cox (RCox)~\cite{glmnet_cox}.
(Section~\ref{sec:MTLR} presented another approach,
based on the more recent
MTLR approach to survival analysis.)

\subsection{\ParamAlg{PreProcess}{}{}\ and \ParamAlg{PreProcess'}{}{}} 
\label{app:PreProcess}

The \PreProcess\ process (used by \LSM\ in Figure~\ref{fig:LSM})
applies various standard ``normalizations'' and simple ``corrections''
to the training data -- both raw clinical features, and gene expression values.
For the clinical features $X_{CF}$, 
\PreProcess\ produces a normalized dataset without any missing values, 
ready for the subsequent steps in the pipeline -- 
see the orange-lines in Figure~\ref{fig:LSM}. 
This uses the standard steps:
(1)~impute missing real (resp., categorical) values for a feature with the mean (resp., mode) of the observed values for that feature;
and
(2)~binarizing each categorical variable (aka ``one-hot encoding'')
-- \eg we encoded the 12-valued ``Histological type'' using twelve bits:
\eg
{\em Invasive Tumor} is [1,~0, 0, 0, 0, 0, 0, 0, 0, 0, 0, 0]. 
For the gene expression data $X_{GE}$, \PreProcess\ applies the following steps: (1)~As we want to deal with the $\log$ of the initial gene expression value,
we first $\log_2$-transformed the data, if necessary. (Below we use ``gene expression'' to refer to this transformed value.) (2)~Then translate all expression values into their ``common z-scores''. 
It first computes the (common) mean and standard deviation over all the genes from the entire $X_{GE}$ dataset: 
Let $\gev{i}{j}$  be the expression value of probe/gene $g_i$ of patient $j$, 
then compute the common mean $\hat{\mu} = \frac{1}{n} \sum_{i,j} \gev{i}{j}$
 (where $n =$ 1,981$\times$49,576 is the total number of entries for METABRIC), and  
the variance $\hat{\sigma}^2 = \frac{1}{(n-1)} \sum_{i,j} (\gev{i}{j}  - \hat{\mu})^2$.
We then use the Z-score transformation of each entry:  
$z_i^j = \frac{(\gev{i}{j} - \hat{\mu})}{\hat{\sigma}}$.
Notes: (a)~this standardization is done prior to dividing the data into train and validation sets. (b)~Using z-scores based on only a single gene
would not be able to identify which genes did not vary much, 
as (after this transform) all genes would vary the same amount.
(3)~\PreProcess{} then removes the genes that do not vary much,
removing a gene $i$ iff its $z_i^j$ values 
are all within the first standard deviation --
\ie if $\forall j\ z_i^j \in [-1, +1]$.

This filtering process is motivated by the assumption that any gene whose expressions does not change much across multiple patients, 
is unlikely to be directly related to the disease, 
while the genes that contribute, typically have significant variations in their expression levels across patients. 
While this filtering procedure is unsupervised,
we anticipated that it would retain the genes that have the most prognostic ability. 
This was confirmed as we found that this process does not eliminate any of the ``top'' $100$ probes in the METABRIC data
(these are the probes with the $100$ highest concordance values); 
see \cite[Table~1]{cheng2013development}.
In METABRIC, this filtering procedure eliminates 27,131 of the original 49,576 probes, 
leaving only 22,445 probes -- 
\ie a $\approx$54.7\% reduction in the number of features.

Later, the performance system 
\ParamAlg{USM}{}{}\ will need to apply these pre-processing steps to a novel instance 
-- in particular, for each clinical feature,
it will need to know the mean (or median) value, for imputation.
Similarly, it will need to transform each gene expression values into an integer;
this requires knowing the global mean $\hat{\mu}$ and $\hat{\sigma}^2$ values
to produce the $z$-values $\{ z_i^j\}$ values,
We include all of these values in the $\Omega$ term, 
which is output by the \PreProcess\ subroutine.
This $\Omega$ is one of the inputs to the \PreProcess{}' process, within \USM,
which applies these pre-processing steps to a novel instance encoded by its $x_{GE}$ and $x_{CF}$ features.
Note finally that neither \PreProcess\ nor \PreProcess{}' use the labels (survival times).

\subsection{ \ParamAlg{ComputeBasis}{[$\rho=$dLDA]}{} } 
\label{sec:ComputeBasis}

As shown by the blue lines in Figure~\ref{fig:LSM},
the \ComputeBasis\ process takes as input a pre-processed version of 
the labeled dataset that was input
to \LSM:
the \PreProcess{}ed gene expression data $X'_{GE}$
and clinical features $X'_{CF}$,
with their associated labels ($\Lbl$).
This process produces 
the ``basis'' set,
of type $\rho$.
This subappendix will focus on $\rho\,=\,$dLDA. 

\begin{algorithm}[t]
{
  \caption{  \ParamAlg{ComputeBasis}{[$\rho$=dLDA]}{}\
  algorithm \label{alg:dLDA}}
  \scriptsize \begin{algorithmic}[1]
   \Function{\ParamAlg{ComputeBasis}{[$\rho=$dLDA]}{( $X'_{GE}, \ X'_{CF},\ Lbl$ )}  }{}
   \Comment{Returns a set of \strain{}s} 
    \State $X''_{GE}$\ :=\ Discretize( $X'_{GE}$ )
      \For{\texttt{t in \{ \tech{A}, \tech{B} \}} } 
        \State \textsf{$GE_t$ := Encode-GE( t, $X''_{GE}$ )}
        \State $[ GE_{t, 1},\ GE_{t, 2} ,\ \ldots, GE_{t, 5} ]$ := \textsf{Partition}( $GE_{t}$ ) 
        \State \texttt{\% Notation: $GE_{t, -i}\ =\ GE_t\ -\ GE_{t, i}$}
      \State \texttt{\% $X'_{CF, i}\ =\ $clinical features;}
        \For{\texttt{K in (5, 10, 15, $\ldots$, 150)}} 
        	\For{\texttt{i = 1:5}} 
            	\State \texttt{\% Find LDA "basis set" (set of \strain{}s)}
        		\State $\wordProb_{t, K, i}$ := \ComputedLDA( $GE_{t, -i}$,\ K )\\
                
                \State \texttt{\small \% Project the hold-out set onto this basis set,
                 }           \State \texttt{\%
                encoding each patient as a K-tuple of values}
                \State $GE\_Topic_{t, K, i}$ := \UsedLDA( $GE_{t, i}$, $\wordProb_{t, K, i}$ )\\
                
                \State \texttt{\% Learn a Cox model for this encoding, value of $K$, and fold $i$}
                \State \texttt{\% using both \strain{}s and the clinical features}
                \State \texttt{\% LearnCox \& PredictCox are based on \cite{coxmodel}}
                \State $w_{t,K,i}$ := 
                  \textsf{LearnCox}( $[\,GE\_Topic_{t, K, i},\ X'_{CF,i}\,],\ Lbl_{i}$ )\\
                
                \State \texttt{\% Evaluate model on the hold out set}
                \State \texttt{\% Using evaluation measures concordance and likelihood}
                \State $c_{t, K, i}$ := \textsf{Concordance}( \textsf{PredictCox}( $w_{t, K, i},\ [GE_{t, i},\ X'_{CF,i}]$), $Lbl_i$ )  
                   
                \State $l_{t, K, i}$ := \textsf{Average}\{ \textsf{Likelihood}( $\wordProb_{t, K ,i}$, $GE_{t, i}$ ) $\}$
                \\
        	\EndFor
            \State $\bar{c}_{t, K}$ := \textsf{Average}$\{\ c_{t, K, i}\ \}$
            \State $\bar{l}_{t, K}$ := \textsf{Average}$\{\ l_{t, K, i}\ \}$
            \State 
        \EndFor
      \EndFor
      \State \texttt{\% Find $t^*$ (encoding scheme), with the highest concordance}
      \State $t^*$ := $arg\max_{t} \{\ \bar{c}_{t, K}\ \}$
      
      \State \texttt{\% Selecting $K^*$} 
      \State $\hat{K} \ = \ arg\max_{K}(\bar{l}_{t^*, K})$
     \State $K^* = \hbox{arg}\max_{K\ s.t.\ 
       \bar{l}_{t^*, \hat{K}} \ - \ 
       \hat{\sigma} 
       (l_{t^*, \hat{K}}) 
        \  \leq \ \ \bar{l}_{t^*, K} }
         \{ \bar{c}_{t^*, K} \}$
     \texttt{\% Break ties giving priority to small K's}
       \\  
      \State \textbf{return} \ComputedLDA( $GE_{t^*},\ K^*$ )
    \EndFunction
  \end{algorithmic}
 }
\end{algorithm}
As shown at the bottom of Algorithm~\ref{alg:dLDA},
\ComputeBasis\ returns the results of
  \ComputedLDA( $GE_{t^*},\, K^*$),
which are a set of $K^*$ \strain{}--distributions,
based on its input $GE_{t^*}$,
which encodes the gene expression values ($X'_{GE}$)
as non-negative integers.
This means \ComputeBasis\ must first 
(1)~transform its input real-valued gene expression values $X'_{GE}$ into non-negative integers $GE_{t^*}$,
and
(2)~determine the appropriate number of \strain{}s
$K^* \in {\cal Z}^+$ .
Task~(1) has two parts:
(1a)~Line~2 first discretizes the real-valued $X_{GE}$
into (positive and negative) integers ${\cal Z}$.
(1b)~The next part of the subroutine
determines the best way to transform those integers 
into {\em non-negative integers}\ ${\cal Z}^{\geq 0}$.
Below we describe these three steps, 
followed by 
(3)~a description of \ComputedLDA.

\subsubsection*{(1a)~{\sc Discretize} subroutine}
Recall first that the \PreProcess\ routine already translated the real-valued $X_{GE}$ gene expression values
into z-scores $X'_{GE}$, 
and excluded every genes whose values here all were in $(-1,\,+1)$.
To simplify the notation, view $X'_{GE} = \{ z_i^j\}$.
The {\sc Discretize} routine first assigns each 
$z_i^j \in\ (-1, 1)$ to $0$.
For the remaining ``non-trivial'' standardized gene expression values $z_i^j$'s 
(outside the first standard deviation) of each gene:
Letting $Z_i^+ \ =\ \{\ z_i^j \ |\ z_i^j \geq 1\ \}$
be the non-trivial positive values, we divide
    $\Delta_i^+\ =\ \max \{ Z^+_i\}\ -\ \min\{ Z^+_i\}$
 into 10 regions, 
 of size $\Delta_i^+/10$ each
 and identify each positive $z_i^j$ with the index $\in\,$\{ 1, 2, \dots, 10\} of the appropriate bin. 
 We similarly divide the non-trivial negative expression values 
 $Z_i^-\ =\ \{ z_i^j \ |\ z_i^j \leq -1\ \}$
 into their 10 bins, based on
  $\Delta_i^-\ =\ \max \{Z^-_i\} - \min \{Z^-_i\}$ and each negative $z_i^j$ is identified with the index $\in$\{ -1, -2, \dots, -10\} of the appropriate bin;
see Figure~\ref{fig:encoding_AB}.
(Of course, the actual divisions are specific to the different genes;
this figure just shows a generic split.)
In general, we let $b_i^j$ be the integer bin index
associated with gene $g_i$ for subject $j$.
\footnote{1. We initially tried to discretize the values into the bins associated with the standard deviation, in general.
However, we found this did not work well.\\
2.~\ComputeBasis{} also returns these 
$\{\Delta_i^+, \Delta_i^-\}_i$ values,
as part of the encoding --
\ie along with $\wordProb_{GE}$ --
and \UseBasis\ will later use this information to discretize
its real-valued gene expression input.
We did not show this detail,
to avoid overcluttering the text and images.
}

\subsubsection*{(1b)~Transform to Non-Negative Integers}
While {\sc Discretize} mapped each gene expression value $z_i^j$ to an integer $b_i^j$,
the \ComputedLDA{} routine requires \em non-negative} values.
Section~\ref{sec:ComputeBasis-dLDA} discussed two ways
to deal with this: 
using either encoding \tech{A}\ versus \tech{B};
see bottom of Figure~\ref{fig:encoding_AB}.
\ComputeBasis\ uses internal cross-validation to determine 
which of these is best, $t^*$, 
along with the number $K^*$ of \strain{}s;
see below.
(In general, we will let $GE_{t}$ refer to the $t$-encoding of 
the gene expression values.)

\subsubsection*{(2)~Finding Optimal $K^*$\add[m]{, $t^*$}}
As noted, the \ComputedLDA\ algorithm also needs to know the number of \strain{}s $K^*$ to produce.
Rather than guess an arbitrary value, 
\ComputeBasis{} instead uses (internal) cross-validation to find the best value for $K$, over the range $K \in \{5,\ 10,\ 15, \dots, 150\}$.
For each technique $t \in \{$\tech{A}, \tech{B}$\}$ and 
each of the 30 values of $K$,
\ComputeBasis\ first computes the dLDA model over the training set,
using \ComputedLDA\ (for that encoding and number of \strain{}s);
it then used these and the (preprocessed) clinical features ($X'_{CF}$) as covariates,
along with the survival labels {\em Lbl},
to learn a Cox model~\cite{coxmodel}
-- see Algorithm~\ref{alg:dLDA},
lines 9--20.
Note it does this in-fold -- 
using 4/5 of the training set to learn the dLDA \strain{}s and the Cox model, 
which is evaluated by computing the concordance (based on this learned model) on the remaining 1/5 (line~24).

As noted above, we need to 
determine 
(1b)~which is the best discretization $t^*$, \tech{A} or \tech{B}, and
(2)~what is the appropriate $K^*$ for that technique.
To answer the first question, 
\ComputeBasis\ 
picked the encoding technique $t^*$ that gave the highest cross-validation concordance from all the ($30\times 2$) combinations 
(see Algorithm~\ref{alg:dLDA},~line 31). 
Secondly, after deciding on a encoding scheme, 
it sets $\hat{K}$ to be the value with the 
largest (cross-validation) likelihood,
then selects the set of $K$'s 
that are smaller than $\hat{K}$ and
whose cross-validation likelihood scores are within the first standard deviation of the $\hat{K}$'s;
see Algorithm~\ref{alg:dLDA}, line 33.
From these candidates, 
it selected the $K^*$ that gives essentially the highest concordance (see Algorithm~\ref{alg:dLDA},~line 34). 
\add[R2]{Empirically, we found that the internal cross-validation concordance scores was fairly flat over the critical region
-- \hbox{\eg} $K \in \{20, .., 35\}$ for METABRIC --
before dipping to smaller values for larger value of K, presumably due to overfitting.  This is why we are confident that the upper limit, of 150 topics, is sufficient.
}
Once it finds the best $K^*$ and the encoding technique $t^*$,
\ComputeBasis\ then runs \ComputedLDA\ 
on the $t^*$-encoded 
(preprocessed) 
training gene expression data $GE_{t^*}$,
seeking $K^*$ \strain{}s; 
this is $\wordProb_{GE}$ ``basis''.
This routine also returns the $\{ \Delta^{\pm}_i\}$
values used to produce the discretized values, $GE_{t}$.

\subsubsection*{ (3)~\ComputedLDA }
The \ComputedLDA( $GE_t,\ K$) process, based on Blei~\etal~\cite{blei2003latent},
computes $K$ \strain{}s, based on the preprocessed, discretized gene expression data $GE_{t}$, 
as well as 
the number of latent \strain{}s $K$;
 it then returns $K$ \strain{}s--distributions,
 each $\approx$50,000-parameters of the Dirichlet distribution
 \add[m]{(for METABRIC)},
 corresponding to a line of the $\wordProb_{GE}$ shown in
 Figure~\ref{fig:C-U-Basis}.
 (Each point here corresponds to its 
 estimate of the posterior $\wordParam_{GE}$, 
 conditioned on the observed gene expression values.)
This routine also uses the Dirichlet prior for the patient--\strain{}s distribution;
here we used the symmetric Dirichlet($\alpha,\, \dots, \alpha$) for some $\alpha \in \Re^{>0}$.
(As there are $K$ \strain{}s; 
we view this as a vector $\alpha {\bf 1}_K$.)
We experimented with several values 
$\alpha\, \in \,\{0.01,\ 0.1,\ 0.5,\ 1.0\}$, 
but found that the prior did not make much difference, 
since we allowed the model to estimate the prior internally. 
We therefore set $\alpha = 0.1$.

 \def\bprior{\beta^{(*)}} 
This routines also needs to set the priors for
the $K$ different
\strain--gene\_expression distributions
$\bprior_{GE}[i,:]= [ \bprior_{GE}[i,1], \dots, \bprior_{GE}[i, N] ]$, for 
$i=1..K$,
each sweeping over the $N$ genes. 
Here, we use the prior
$\bprior_{GE}[i, j] \ = \frac{1}{N}+\delta$
where $\delta \ \sim U[0,\,1/N^2]$
-- \ie $\delta$ is sampled from the uniform distribution over the interval 
$[0,\ 1/N^2]$. 
This LDA learning process~{\cite{blei2003latent}}
uses the data in $GE_t$ to compute the posterior distribution
 $\{\,\beta_{GE}[i,:]\,\}_i$ for each of these $K$ \hbox{\strain}s
 -- revealing $GE_t$'s intrinsic structure.
Recall these are just the parameters for Dirichlet distribution; 
note they must be positive, but do not add up to 1.
The \ComputedLDA\ returns the expected values of the gene expression values drawn from this posterior distribution:
{\large $\wordProb_{GE}[i,j] = \frac{\wordParam_{GE}[i,j]}{\sum_{j'}\wordParam_{GE}[i,j']}
$}.
Here, the probability values for each \strain{} $\wordProb_{GE}[i,:]$
add up to 1.
We will let $\wordProb_{GE} = \{ \wordProb_{GE}[i,:]\}$ refer to the entire ``matrix''.

\subsection{ \ParamAlg{UseBasis}{[$\rho=$dLDA]}{} }  
\label{sec:UseBasis}

Once LSM has learned the \strain{}s 
($\wordProb_{GE}$) 
for the best $K^*$ and best encoding technique $t^*$, 
we can then compute the \strain{} distribution for a new patient (based on her gene expression $x_{GE}$);
see Figures~\ref{fig:LSM} and~\ref{fig:usm_workflow}.
This will call \UseBasis,
which in turn runs  \UsedLDA\ 
(the {\em LDA inference procedure}) 
on the preprocessed gene expression data $x'_{GE}$ 
of the current patient
to compute the individual 
topic contributions for this patient~\cite{blei2003latent}. 
The inference procedure determines the
posterior distribution of the {\bf patient-\strain} Dirichlet distribution
$\Theta(x'_{GE})\ =\ 
[\,\docParam_1(x'_{GE}),\
\docParam_2(x'_{GE}),\ \dots,\ 
\docParam_{K^*}(x'_{GE})\,]\ \in\ \Re_+^{K^*}$,
where each $\docParam_j(x'_{GE})\ \in\ \Re_+$ 
quantifies how much of this patient's gene expression is from the $j^{th}$ \strain\
(using the posterior mean probabilities of the
\strain{}s, $\wordProb_{GE}$).

This process reduces the
$\approx$20\,000-dimension gene expression values to a very small $K^*$-dimensional
\strain{}s representation
-- \eg $L^* = 30$.
These low-dimensional feature vectors are then used in the survival prediction algorithms to predict survival times/risk.

\subsection{\ParamAlg{ComputeBasis}{[$\rho=$PCA]}{}
 and \ParamAlg{UseBasis}{[$\rho=$PCA]}{}  } 
\label{sec:superpc+}
The previous 
subappendix described one way to reduce the dimensionality of the data --
to transform each patient's 
20,000-tuple
to a more manageable $K$-tuple
-- there based on topic modeling ideas. 
There have been many other feature selection methods proposed for survival prediction using gene expression data, 
such as hierarchical clustering, univariate gene selection, supervised PCA, penalized Cox regression and tree-based ensemble methods~\cite{van2009survival}.
Some of these techniques first apply a procedure to reduce the dimensionality of the data, based on feature selection, feature extraction or a combination of both, 
while others, such as random survival forests~\cite{RandomSurvivalForests} and L1-penalized Cox~\cite{goeman2010l1}, include internal feature selection.
As we wanted to compare our dLDA approach to other dimensionality reduction techniques,
we chose an extension to the principle component analysis called supervised principal component analysis (SuperPC)~\cite{bair2004semi}, instead of other regularization techniques.

This algorithm first calculates the univariate Cox score statistic 
of each individual gene against the survival time,
then retains just the subset of genes whose score exceeds a threshold, 
determined by internal cross-validation. 
 Then it computes PCA on the dataset containing only those selected genes,
 then projects each patient onto the first one (or two) components.
The main disadvantage of the SuperPC algorithm is that the individual genes 
selected from the univariate selection process
might not perform the best in a multivariate (final) model,
perhaps because many of these top-ranked genes may be highly correlated with one another --
\ie it would be better having a more ``diverse'' set of genes~\cite{van2009survival,ding2005minimum}.
Instead, we use a variant, called SuperPC+, 
that initially applies PCA on the {\em normalized} gene expression data after the constant genes are removed;
see \PreProcess\ in Appendix~\ref{app:PreProcess}).
The PCA transformation 
projects the initial ``raw'' features into a different space, which then can be used to select the top components based on the univariate Cox regression.
Note this SuperPC+ is (still) computationally efficient, as it is based on PCA, which is efficient:
Even though gene expression data is high dimensional
($p \gg n$, where $p$ is the number of genes and $n$ is the number of instances),
the rank of the GE matrix will be (at most) $\min\{p,n\} = n$.
Therefore, PCA can be performed without many computational restraints on the whole gene expression dataset, as here the PCA time complexity is $O(n^3)$.
After performing PCA on the GE dataset, we can then identify the most important principal components by computing a Cox score statistic for the univariate association between each principal component and the survival time. 
In our experiments, we select the threshold $\eta$ for the p-value of the Cox score by internal cross-validation (wrt concordance), and retained all PCs having a p-value lower than this $\eta$
-- finding $\eta=$5e-4 for the METABRIC dataset and $\eta=$5e-2 for KIPAN.
These selected PC components 
form the basis set $B_{GE}[\rho=$PCA].

\ParamAlg{UseBasis}{[$\rho=$PCA]}{} 
is simply the projection of the gene expression data into the chosen PC components.
This gives us a low dimensional feature representation of the original gene expression data to feed into the survival prediction algorithms.

\subsection{ Cox Models: \ParamAlg{LearnModel}{[$\Psi=$Cox]}{},
\ParamAlg{LearnModel}{[$\Psi=$RCox]}{} } 
\label{sup:cox}
The Cox regression model's~\cite{coxmodel} hazard function over time $t$,
for an individual described by $x$,
is the product of two components:
\begin{equation}
 h_W(\,t\,|\,x\,) \quad = \quad h_0(t) \times \exp(x^T W)
 \label{eqn:HR-Cox}
\end{equation}
where the baseline hazard $h_0(t)$ is independent of the covariates $x$
and the covariates 
are (independently) multiplicatively related to the hazard,
based on a (learned) $W$. 
This formulation simplifies modeling of the hazard function by limiting the contribution of the ``time'' variable $t$ to the baseline hazard $h_0(\cdot)$, which means
the hazard ratio (HR) between two patients
\begin{equation*}
\hbox{HR}(x_1,\, x_2) \ = \ \frac{h(\,t\,|\,x_1\,) } {h(\,t\,|\,x_2\,)} \ = \ 
\frac{h_0(t) \exp(x_1^T W)} {h_0(t) \exp(x_2^T W)} \ = \ \exp((x_1 - x_2)^T W)
\label{eqn:Lc-Cox} 
\end{equation*}
does not depend on time and is linear (proportional) in the exponent.
To estimate the coefficients of the model, 
Cox~\cite{coxmodel} proposed a partial likelihood technique that eliminates the need to estimate the baseline hazard.
This procedure allows the Cox proportional hazards model to be semi-parametric,
by only using the survival times to rank the patients~\cite{steck2008ranking}. 
We can compute partial likelihood with all patients -- both censored and uncensored:

\begin{equation}
L_c(W) \quad = \quad 
\prod_{i=1}^{N} \left(
\frac{\exp({x_{i}}^{T} W)} 
{\sum_{k \in \mathcal{R}(y_{i})} \exp({x_{k}}^{T} W)}
\right )^{\delta_i}
\label{eq:cox}
\end{equation}
\begin{itemize}
\item \change[R1]{$\mathcal{R}(r_j)$}{$\mathcal{R}(y_j)$} is the risk set at time $y_j$, which are the indices of individuals who are alive and not censored before time $y_j$
\item \change[R1]{$[x_i, r_i, \delta_i]$}{$[x_i, y_i, \delta_i]$} describes the $i^{th}$ subject, where\\
\hspace*{1em} $x_i$ = vector of covariates\\
\hspace*{1em} \change[R1]{$r_i$}{$y_i$} = (survival or censor) time\\
\hspace*{1em} $\delta_i$ = censor bit (0 for censored; otherwise 1)
\item $N$ -- total number of patients in the cohort
\item $W$ -- coefficients (to be learned)
\end{itemize}

Note that only the uncensored likelihoods contribute directly,
since for censored instances $\delta_i \, = \, 0$.
Therefore the censored observations are only utilized in the denominator when summed over the instances in a risk set. 
In essence,
the partial likelihood only uses the patient's death times to rank them in the ascending order to find the risk sets and does not use the exact times explicitly~\cite{steck2008ranking}. 
Hence, the coefficients estimated by maximizing the partial likelihood depend only on the ordering of the patient's death times 
and the covariates,
allowing for an implicit optimization for good concordance of the risk score. 
An in-depth study on the Cox proportional hazard model has revealed that the partial likelihood proposed by \cite{coxmodel} is approximately equivalent to optimizing concordance~\cite{steck2008ranking}.

There are several extensions of the basic Cox proportional hazards model:
some extend the initial model estimating the baseline hazard and others are based on the regularization methods imposed on the coefficients ($W$). 
Generally, regularization based on LASSO, ridge penalty or the elastic-net regularization (which allows both L1 and L2 penalties) are adopted to reduce overfitting. 
In our work,
we use the {\tt glmnet} R package~\cite{glmnet_cox} with ridge penalty
(by setting $\alpha=0$ in the glmnet function); 
here called RCox. 
We selected ridge penalty based on the internal cross validation.
We found that concordance results using models with ridge penalty were better than those having no regularization (LASSO, elastic-net).

\section{Foundations} 
\label{app:Found} 

\subsection{Evaluation: Concordance Index (CI)} 
\label{app:CI}
This ``CI'' evaluation applies to any model that assigns a real number -- a ``risk score'' -- to each instance $f(\cdot)$.
It considers all pairs of ``comparable'' instances, and determines
which is predicted 
(by the risk model $f(\cdot)$)
to die first,
and also who actually died first.
CI is the proportion (probability) of these pairs of instances whose actual pair-wise survival ordering,
matches the predicted ordering, 
with respect to
$f(\cdot)$:
\begin{equation}
\label{eq:CI}
\hbox{CI}( f ) \quad = \quad \frac{1}{|\Psi|} \sum_{(i,j) \in \Psi} I[\,{f(x_i) > f(x_j)}\,]
\end{equation}
where $I[\phi]$ is the indicator function,
which is $1$ if the proposition $\phi$ is true, and $0$ otherwise.
A pair of patients is ``comparable'' if we can determine which died first --
\ie if both are uncensored, 
or when one patient is censored after the observed death time of the other;
this corresponds to the set of 
ordered pairs of indices $\Psi$.
This CI($f$) score is a real value between 0 to 1, where 1 means all comparable pairs are predicted correctly. 
CI can be viewed as a general form of the Mann--Whitne--Wilcoxon statistic and is similar to 
the Area Under the 
ROC (Receiver Operator Curve), AUC,
of classification problems~\cite{steck2008ranking}.

\subsection{Evaluation: D-calibration} 
\label{app:d-calib}
The concordance index is a discriminatory measure,
which is relevant, for example, 
when deciding
which patient with liver failure will die first 
without a transplant.
By contrast, {\em calibration}\ measures the deviation between the observed and the predicted event time distributions.
While this is not meaningful if we only have a risk score
(\eg as produced by the basic Cox Proportional Hazard function),
this deviation can be computed for a {\em survival distribution},
like ones produced by the MTLR survival prediction tool, or the Cox+KF system
-- which extends the standard Cox model by using the Kalbfleisch-Prentice estimator to produce the baseline hazard function
$h_0(x)$ in Equation~\ref{eqn:HR-Cox}; see \cite{kalbfleisch2011statistical}.
{In general, this
calibration involves computing the difference between the predicted versus observed probabilities in various subgroups --
\eg if the predicted probability of surviving at least $t=$2576~days is 0.75 
for some subgroup,
then we expect to observe around $75\%$ of these patients to be alive at this time $t$.}

\begin{figure}
\centering
\includegraphics[width=5in,height=2in]{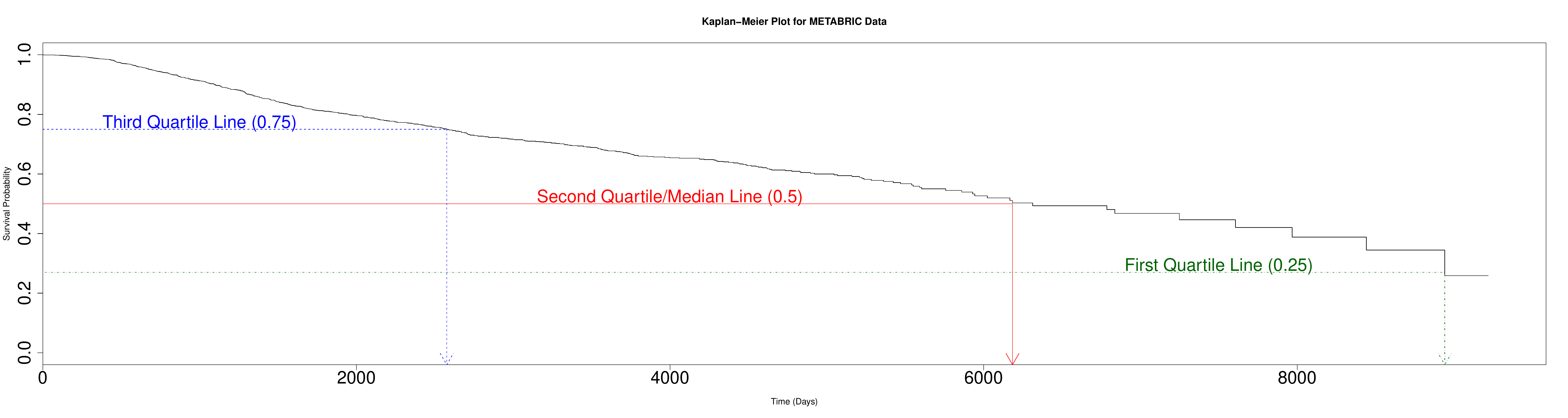}
\caption{Kaplan--Meier survival function from METABRIC (training) data. (We only use the quartiles for pedagogic purposes)}
\label{fig:km_plot}
\end{figure}

We consider a novel measure of the calibration of such survival curves, called D-calibration
(``D'' for ``Distribution'')~\cite{ISD-Paper}.
To motivate this, consider a standard Kaplan-Meier (KM)~\cite{kaplan-meier} 
plot shown in Figure~\ref{fig:km_plot},
which 
plots the set of points $(t,\ \KM{t}\,)$ -- 
\ie it predicts that the $\KM{t}$ fraction of patients will be alive at each time $t \geq 0$.
Hence, the point (6184~days, 0.50) means the median survival time of the cohort is 6184~days;
see Figure~\ref{fig:km_plot}(solid line).
We will use $\KMinv{p}$ to be the time associated with the probability $p$ -- technically, 
$\KMinv{p}$ is the earliest time when the KM curve hits $p$;
hence $\KMinv{0.5} =\ $6184~days.
If this plot is D-calibrated, then around 50\% of the patients
(from a hold-out set, not used to produce the KM curve)
will be alive at this median time.
So if we (for now) ignore censored patients,
and let $d_i$ be the time when the $i^{th}$ patient died, 
 consider the $n$ values of $\{\, \KM{d_i}\, \}_{i=1..n}$.
 Here, we expect $\KM{d_i} > 0.5$ for 1/2 of the patients.
Similarly, as the curve includes (2576~days,\ 0.75) and 
(8941~days,\ 0.25),
then we expect 75\% to be alive at 2576~days, and 25\% at 8941~days;
see Figure~\ref{fig:km_plot}.
Collectively, this means we expect 25\% of the patients to die between $\KMinv{1.0}\ =\ $0~days and $\KMinv{0.75}\ =\ $2576~days, 
and another 25\% between $\KMinv{0.75}$ and $\KMinv{0.5}$,
etc.
These are the predictions; we can also check, to see how many people actually died in each interval: 
in the first quartile (between 0 and 2576~days), in the second (between 2576 and 6184~days), in the third (between 6184 and 8941~days), and the fourth (after 8941~days).
If the KM plot is ``correct'' -- \ie is D-calibrated --
then we expect 1/4 of the patients will die in each of these 4 intervals.
The argument above means we expect 1/4 of the $\{\, \KM{d_i}\, \}$ values to be in the interval [0, 0.25], and another quarter to be in [0.25, 0.5], etc.
Stated more precisely, 
\begin{equation}
\hbox{the values of 
$\{\, \KM{d_i}\, \}$ are uniformly distributed.} 
\label{eqn:KM}
\end{equation}

A single KM curve is designed to represent a cohort of many patients.
The MTLR system, however, computes a different survival curve for each patient -- call it $Pr_i(\,\cdot\,)\ =\ \PSSP{W}{\cdot}{\pat_i}$ (from Equation~\ref{eqn:PSSP}).
But the same ideas still apply: 
Each of these patients has a median predicted survival time -- 
the time $Pr_i^{-1}(0.5)$ where its $Pr_i(\,\cdot\,)$ curve crosses 0.50. 

By the same argument suggested above, 
we expect (for a good model $W$) 
that 1/2 of patients will die before their respective median survival time --
$d_i \leq Pr_i^{-1}(0.5)$;
that is, $| \{ i\, :\, Pr_i(d_i) \leq 0.5 \} | \approx n/2$.
Continuing the arguments from above, we therefore expect
the obvious analogue to Equation~\ref{eqn:KM}: 
\begin{equation}
\hbox{the values of 
$\{ Pr_i(\,d_i\,) \}$ are uniform.} 
\label{eqn:PiDi}
\end{equation}

We can now test whether a model is D-calibrated by using the Hosmer-Lemeshow (HL)~\cite{hosmer2013applied} 
goodness-of-fit test,
which compares the difference between the predicted and observed events in the event subgroups:
\begin{equation}
\hbox{HL}\left(\
\left[  
{\begin{array}{c}
[N_1,\ P_1,\ E_1]\\
 \cdots\\ {}
[N_G,\ P_G,\ E_G]\\
\end{array}}
\right]
\right)\quad = \quad  
\sum_{g=1}^{G} \frac{ (E_{g} - P_{g})^2 } {N_{g}\pi_g (1 - \pi_g)}
\label{eq:hl}
\end{equation}
where $G$ is the number of subgroups (here 4),
where the $g^{th}$ subgroup has 
$N_{g} \in \mathbb{Z}^+$ events,
with the empirical number of events $E_{g} \in \mathbb{Z}^+$ 
(which here is $N/g$, for $N =\sum_g N_g$ total patients), 
the corresponding predicted number of events 
in each group
$P_{g}\in \mathbb{Z}^{\geq 0}$,
and $\pi_{g} = \frac{N_g}{N}$
(which here is $\frac{1}{G}$)
is the proportion of the $g^{th}$ subgroup.
Under the null hypothesis (Equation~\ref{eqn:PiDi}), this $\hbox{HL}$ statistics follows a Chi-Square distribution
with $G-2$ degrees of freedom.
If the predicted and empirical event rates are similar 
for the subgroups, the test statistic will fail to reject the null hypothesis,
providing evidence that the model's predictions are well D-calibrated 
-- \ie large p-values from the test statistic 
suggest not rejecting the null hypothesis).

Notes: (1)~This evaluation criterion only applies to models that produce survival distributions,
which means it directly applies to the MTLR models.
For the Cox and RCox models, we used the 
Kalbfleisch-Prentice baseline hazard estimator~\cite{kalbfleisch2011statistical} to produce personalized survival curves.
(2)~To provide more precise evaluation, rather than using 4 bins (quantiles), we mapped the 
$Pr_i(\,d_i\,)$ probabilities
into 20 bins: 
[0, 0.05); [0.05, 0.1), \dots, [0.95, 1.0].
(3)~This analysis deals only with uncensored data; 
Haider~\etal~\cite{ISD-Paper}
discusses how to cope with censored data.

\section{Additional Results} 
\label{app:Results}

This appendix presents additional results:
First, Appendix~\ref{sup:lpd}
evaluates the Latent process decomposition (LPD) method,
then Appendix~\ref{sup:calibration} provides D-calibration results of our various models.

\subsection{Latent process decomposition (LPD) for microarray feature extraction} 
\label{sup:lpd}
Rogers~\etal~\cite{rogers2005latent} introduced LPD as a topic model adaptation for microarray data.
We experimented with LPD (on METABRIC data) to derive genetic features and used them along with the clinical features for comparison. 
We used internal cross-validation for LPD to find
the optimal number of latent processes
for the METABRIC data --
and found that 10 was best.

We then used the model based on these 10 latent process;
the resulting concordance results,
on the hold-out dataset,
was 0.6915 (Cox), 0.6077 (RCox) and 0.6995 (MTLR).
Comparing this to the ``B'' and ``C'' rows of 
Table~\hbox{\ref{tab:metabric}},
we see that our dLDA approach performs better than this complex adaptation of the LDA model for microarray data,
for the survival prediction task
-- \ie dLDA 
produces better features from the gene expression data.

There are two other reasons to prefer our dLDA-approach:
(1)~LPD has large time and memory requirements. 
(2)~Moreover as our dLDA directly uses the LDA model,
it can utilize all available off-the-shelf implementations, across several technology platforms with efficient and scalable implementation~\cite{hoffman2010online}.

\subsection{D-Calibration Results} 
\label{sup:calibration}

Table~\ref{tab:calib} shows the D-calibration results
for all of the domain-independent experiments we ran
-- \ie excluding the ``E'' and ``F'' rows from Table~\ref{tab:metabric}, 
which used features that were specific to breast cancer.
We see that the results were D-calibrated
(\ie had a HL p-value $> 0.05$)
in all 12 situations, for METABRIC and KIPAN
-- for all feature groups  \{ A, B, C, D \},
and all 3 learning algorithms \{ Cox, RCox, MTLR \}.
We note that we found that Cox failed this test
on other datasets, including BRCA. 

\begin{table*}[h!]
\def\sCox{\hspace*{-1em}Cox\hspace*{3em}}
\def\sRCox{\hspace*{0em} RCox\hspace*{1em}}
\def\sMTLR{\hspace*{2em} MTLR}

\begin{adjustwidth}{-2.0in}{0in} 
\centering 
\caption{D-calibration results from METABRIC, KIPAN, and BRCA, on the held-out test data. 
p-values greater than 0.05 suggest the model is good (``D-calibrated''). 
Note that Cox models fail D-calibration test for all feature combinations, for BRCA dataset.}
\label{tab:calib}
\begin{tabular}{@{}l|ccc|c|rr@{}|rr@{}|rr@{}}
\hline
\rowcolor{gray!20}
ID &  \multicolumn{3}{c|}{Feature Groups}
   & Algorithms & \multicolumn{2}{c|}{KIPAN} & \multicolumn{2}{c|}{METABRIC} & \multicolumn{2}{c}{BRCA}\\ 
 \rowcolor{gray!20}\tiny
 & \tiny Clinical & \tiny PCA & \tiny dLDA & & \tiny HL-Statistic &\tiny p-value &\tiny HL-Statistic & \tiny p-value &\tiny  HL-Statistic &\tiny p-value \\ 
\thickhline
A & \pp & \mm & \mm &\sCox  &  7.1455 & 0.9888 & 12.6765 & 0.8104 & 52.5289& {\bf 3.1e-05} \\ 
A  & \pp & \mm & \mm & \sRCox  & 7.1279 & 0.9890 & 14.8143 & 0.6747 & 4.8816 & 0.9990 \\ 
A  & \pp & \mm & \mm & \sMTLR  & 8.2421 & 0.9748 & 11.2300 &  0.8843 & 6.6300 & 0.9929\\ 
B  & \pp & \pp & \mm & \sCox  & 5.6697 & 0.9973 & 8.1723& 0.9760 & 63.7983 & {\bf 4.9e-07} \\ 
B  & \pp & \pp & \mm & \sRCox  &  2.5769 & 0.9973 & 6.4499& 0.9940 & 6.0013& 0.9962 \\ 
B  & \pp & \pp & \mm & \sMTLR  & 8.4315 & 0.9714 & 10.8421& 0.9009 & 8.9895& 0.9600 \\ 
C  & \pp & \mm & \pp & \sCox  &  10.54166 & 0.9127 & 6.0066& 0.9962 & 48.9904 & {\bf 0.0001} \\
C  & \pp & \mm & \pp & \sRCox &  7.9878 & 0.9993 & 4.1344& 0.9997 & 4.6031& 0.9993 \\ 
C  & \pp & \mm & \pp & \sMTLR  & 6.5578 & 0.9933 &  8.2000& 0.9755 & 9.2421& 0.9539\\ 
D  & \pp & \pp & \pp & \sCox & 14.1951 &  0.7162 & 11.2333 & 0.8842 & 25.2241 & {\bf 0.1189} \\ 
D  & \pp & \pp & \pp & \sRCox  & 8.2420 & 0.9748 &  3.4981& 0.9999 & 6.0813& 0.9959 \\ 
D  & \pp & \pp & \pp & \sMTLR  & 7.4947 & 0.9852 & 6.5158& 0.9936 & 9.0211 & 0.9593 \\ 
\hline 
\end{tabular} 
\end{adjustwidth}
\end{table*}


\section*{Acknowledgments}
We greatly acknowledge Compute Canada for providing us with computing resources. 
We also thank Dream Challenges and Broad Institute for making the METABRIC and
KIPAN cancer data sets publicly available. 

\section*{Funding}
This work has been supported by the funding from Alberta Machine Intelligence Institute (Amii), and from NSERC.

\nolinenumbers
\bibliography{plos_topic_survival}

\end{document}